\documentclass[runningheads]{llncs}
\usepackage{graphicx}
\usepackage{stackengine}
\usepackage{tikz}
\usepackage{comment}
\usepackage{amsmath,amssymb} 
\usepackage{color}
\usepackage{enumitem}

\usepackage{subcaption}
\usepackage{booktabs}

\usepackage{hyperref}
\usepackage[symbol]{footmisc}

\usepackage[accsupp]{axessibility}  

\newcommand{\new}[1]{\textcolor{black}{#1}}

\begin{document}

\pagestyle{headings}
\mainmatter
\def\ECCVSubNumber{2581}  

\title{Studying Bias in GANs \\
through the Lens of Race}


\titlerunning{Studying Bias in GANs through the Lens of Race}
%
\author{Vongani H. Maluleke*
\and
Neerja Thakkar*
\and
Tim Brooks 
\and 
Ethan Weber 
\and
Trevor Darrell
\and
Alexei A. Efros
\and
Angjoo Kanazawa
\and
Devin Guillory}

\authorrunning{V.H. Maluleke, N. Thakkar, et al.}

\institute{UC Berkeley}

\maketitle

\footnotetext[1]{Equal contribution in alphabetical order.}

\begin{abstract}
In this work, we study how the performance and evaluation of generative image models are impacted by the racial composition of their training datasets. By examining and controlling the racial distributions in various training datasets, we are able to observe the impacts of different training distributions on generated image quality and the racial distributions of the generated images. Our results show that the racial compositions of generated images successfully preserve that of the training data. However, we observe that truncation, a technique used to generate higher quality images during inference, exacerbates racial imbalances in the data. Lastly, when examining the relationship between image quality and race, we find that the highest perceived visual quality images of a given race come from a distribution where that race is well-represented, and that annotators consistently prefer generated images of white people over those of Black people. 
\end{abstract}

\section{Introduction}
The computer vision community has wrestled with problems of bias for decades \cite{ponce2006dataset,torralba2011unbiased}. 
As vision algorithms are starting to become practically useful in the real world, this issue of bias has manifested as a serious problem in society \cite{buolamwini2018gender,o'neil_2018,benjamin2019race,raji2019actionable}. 
In particular, GANs \cite{goodfellow2014generative} have significantly increased in quality and popularity over the past few years \cite{brock2018large,karras2019style}, and these models have been shown to contain racial biases \cite{jain2020imperfect,aigahaku,jain2018imagining}. 
As GANs are increasingly used for synthetic data generation and creative applications, there is the potential for racial bias to propagate to downstream applications, and the need for an understanding of the cause of biased outputs. 
In generative image models, the question of whether the source of biased outputs comes from the data with which models are trained (data distribution bias) or the algorithms themselves (algorithmic bias) is unanswered.

In this work, we aim to understand the source of bias in GANs in the context of \new{perceived} race\footnote[2]
{\new{We do not not objectively evaluate the underlying actual race, but rather measure the perceived race of the image. This is because race is a complex social construct and it is not sufficient to evaluate race with only visual features. See section ~\ref{subsec:race}
}}, i.e., can dataset imbalance alone sufficiently describe issues of racial representation in generative image models? \new{Or do algorithmic choices also contribute \cite{HOOKER2021100241}? We consider the following types of bias in generative image models as they pertain to class distributions and image quality: 1) \textit{Data distribution bias}: imbalances in training data that are replicated in the generated data, 2) \textit{Symmetric algorithmic bias}: imbalances in training data that are exacerbated in the generated data, irrespective of which race labels are over or under-represented in the data, and 3) \textit{Asymmetric algorithmic bias}: unequal effects on different classes, dependent on or independent of class representation in the training data.}

We conduct a systematic study, exploring the following research questions:
\begin{enumerate}
    \item Will a racially imbalanced training dataset lead to an even more imbalanced generated dataset?
    \item Will improving sample quality using the commonly employed ``truncation trick" exacerbate an underlying racial imbalance? 
    \item If a generator is trained on an imbalanced dataset, will perceived visual quality of the generated images change depending on class representation? 
\end{enumerate}

We explore these research questions in the context of StyleGAN2-ADA \cite{karras2020analyzing} trained to generate human faces. To measure the impact of dataset imbalance, we first label a subset of FFHQ \cite{karras2019style}, to understand the racial composition of this popular and representative dataset for training generative models of face images.
We also train StyleGAN2-ADA on three datasets with varying controlled ratios of \new{images of persons perceived as Black or white}.
We then measure the \new{perceived} racial distribution of training and generated data, with and without truncation, and study the relationship between quality and race class label distribution. 
To obtain the \new{perceived} racial distribution, we use Amazon Mechanical Turk Annotation (AMT) annotations, as well as a classifier that is calibrated against human performance. The AMT annotations are also used to measure the perceived visual quality of real and generated images.

Our findings show that 1) GANs appear to preserve the racial composition of training data, even for imbalanced datasets, \new{exhibiting data distribution bias} 2) however, truncation exacerbates discrepancies in the racial distribution \new{equally amongst race class labels, exhibiting symmetric algorithmic bias}, and 3) when ranking images by quality, we find that generated images of a given \new{perceived} race are of higher perceived quality when they come from a generator that is over-represented for \new{images labeled as white}, while \new{images labeled as Black} retain constant quality regardless of the training data's racial composition. We also find that both real and generated white
\new{labeled facial images} are consistently annotated as higher quality than \new{real and generated} \new{images of Black people}. \new{It is unclear whether this observed asymmetric algorithmic bias is caused by StyleGAN2-ADA, our human-centric system of evaluation, underlying qualitative discrepancies in training data, or a combination thereof.}

\section{Related Work}

\paragraph{Racial Bias in Computer Vision}

Machine learning models and their applications have a well-documented history of racial bias, spanning vision \cite{buolamwini2018gender,Klare2012}, language \cite{brunet2019understanding,brown2020language,radford2021learning}, and predictive algorithms that have a heavy impact on real peoples' lives \cite{o'neil_2018,noble_2018,Jeff2016COMPAS}.  Numerous research efforts have aimed to evaluate, understand, and mitigate bias, particularly in computer vision. Buolamwini {\em et al.} \cite{buolamwini2018gender} analyzed three automated facial analysis algorithms, and found that all classifiers performed worse on images of individuals with a darker skin type compared to counterparts with a lighter skin type \cite{buolamwini2018gender}. A similar conclusion was made in earlier research by Klare {\em et al.} \cite{Klare2012}, who found that face recognition algorithms consistently performed poorly on young Black females. Phillip {\em et al.} \cite{Phillips2011} showed that machine learning algorithms suffer from the ``other race effect" (humans recognize faces of people from their race more accurately compared to faces of other races) \cite{Phillips2011}. 

\paragraph{Racial Bias in Generative Models}
Image generation models have been shown to contain racial biases \cite{jain2020imperfect,aigahaku,jain2018imagining}. 
AI Gahaku \cite{aigahaku}, an AI art generator that turns user-submitted photos into Renaissance-style paintings, often turns photos of people of color into paintings that depict white people.
The Face Depixelizer, a tool based on PULSEGAN \cite{Sachit2020}, which super-resolves a low-resolution face image, also tends to generate an image of a white person, regardless of input race.
Jain {\em et al.} \cite{jain2020imperfect} demonstrated that popular GAN models exacerbate biases along the axes of gender and skin tone when given a skewed distribution of faces; for example, Snapchat’s beautification face filter lightens skin tones of people of color and morphs their faces to have euro-centric features. GANs can inherit, reinforce and even exacerbate biases when generating synthetic data \cite{jain2018imagining}. 

\paragraph{Racial Bias Mitigation}

Several works have proposed ways to mitigate racial bias in facial recognition systems by modifying models directly or with data sampling strategies \cite{Gwilliam_2021_ICCV,Wang2019_ICCV,zeyu2019}. To reduce racial bias via model modification, Wang {\em et al.} \cite{Wang2019_ICCV} proposed a deep information maximization adaptation network (IMAN), with white \new{faces} as the source domain and other races as target domains. Gwilliam {\em et al.} \cite{Gwilliam_2021_ICCV} performed facial recognition experiments by manipulating the race data distribution to understand and mitigate racial bias. Their work demonstrated that skewing the training distribution with a majority of African \new{labeled images} tends to mitigate racial bias better than balanced training data set. 

\paragraph{Generative Adversarial Networks}

GANs \cite{Goodfellow2016GenerativeAN}, a class of implicit generative models, learn to generate data samples by optimizing a minimax objective between discriminator and generator networks. The discriminator is tasked with differentiating training images from generated images, and the generator aims to fool the discriminator. Modern GANs \cite{brock2018large,karras2019style,karras2020analyzing} are capable of producing high quality images and are increasingly leveraged for image manipulation tasks \cite{ganspace,abdal2020image2stylegan}.

\paragraph{GAN Truncation}
The ``truncation trick" introduced by Brock {\em et al.} \cite{brock2018large} is a sampling technique that allows deliberate control of the trade-off between variety and fidelity in GAN models. At the loss of some diversity, the fidelity of generated images can be improved by sampling from a shrunk or truncated distribution \cite{ackley1985learning,marchesi2017megapixel,brock2018large}. StyleGAN implements truncation by interpolating towards the mean intermediate latent vector in $W$ space \cite{karras2019style}. \new{In this work we} evaluate the impact of truncation on racial diversity in images generated with StyleGAN2-ADA \cite{karras2020training}.

\paragraph{GAN Mode Collapse}

GANs are known to exhibit mode collapse or mode dropping, where certain features present in the training dataset distribution are missing from the distribution of generated images \cite{goodfellow2017nips}.
Many works propose solutions to address mode collapse, such as Wasserstein GANs \cite{pmlr-v70-arjovsky17a,gulrajani2017improved}, Prescribed GANs \cite{dieng2019prescribed}, and Mode Seeking GANs \cite{Mao_2019_CVPR}. In spite of these works, mode dropping is not fully understood. Arora {\em et al.} \cite{arora2017gans} show that the generated distribution of a GAN has a relatively low support size (diversity of images) compared to the training distribution. In the work ``Seeing What a GAN Cannot Generate," Bau {\em et al.} \cite{bau2019seeing} visualized regions of images that GANs are unable to reproduce. They found that higher-quality GANs better match the dataset in terms of the distribution of pixel area belonging to each segmentation class, and that certain classes, such as people, were particularly challenging for the GANs. These works indicate that GANs may exacerbate bias in the training data distribution by dropping certain features or classes in generated images --- in this work, we also analyze whether these effects occur regarding racial bias.

\section{Methodology}


\subsection{Racial Categorizations}
\label{subsec:race}

\new{Race is a dynamic and complex social construct. People can be of multiple races and perception of race is heavily dependent on culture and geography with different meanings and interpretations. As such, all discussion in this work pertains to perceived race by annotators. Despite the complexities and subjectivity involved in analyzing racial perceptions, we choose to study the bias in GANs through the lens of race as this is a topic of societal consequence \cite{jain2020imperfect,aigahaku,jain2018imagining}}.

\new{The decision to use perceived racial classifications over skin color \textit{(tone/shade)} estimates and categorizations (such as Individual Typology Angle (ITA) and Fitzpatrick skin phototype \cite{wilkes2015fitzpatrick}) was driven by the notion that perceived racial categorization is an informative lens through which to study racial bias \cite{hanna2019}, and the availability of the FairFace dataset, a large-scale face dataset with seven race categorizations.
Furthermore, Karkkainen {\em et al.} \cite{karkkainen2021fairface} found that solely relying on skin color/ITA in the FairFace dataset is not sufficient to distinguish race categories.
We condense the seven race categories into three: Black, white, and Non-Black or Non-white, to further reduce perceptual ambiguity (see sec.~\ref{sec:annotation_consistency}). This study does not aim to minimize the importance of understanding these questions in Non-Black or Non-white races, but as a first study we simplify the categorization.}

\subsection{Datasets}
\label{subsec:real-img-data}

This section describes the FFHQ and FairFace datasets. It also explains how we use the datasets: we quantify the racial distribution of FFHQ, train generative models on both datasets to answer our three research questions, and also use FairFace to train a classifier \new{on perceived race}. 

\begin{figure*}[t]
\centering
\includegraphics[width=\linewidth]{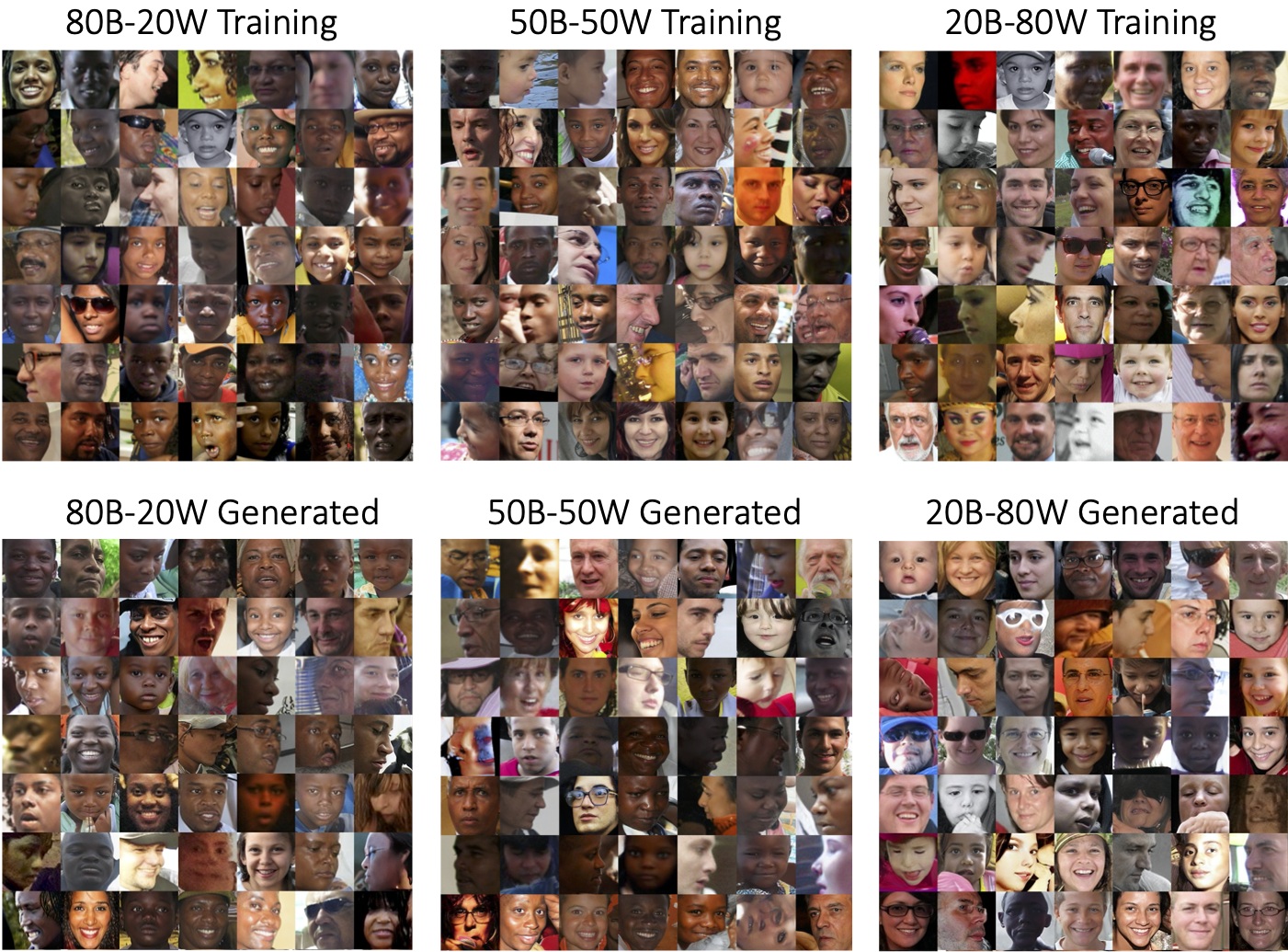}
\caption{\textbf{Training Data and Generated Data.} The top row shows real data from the FairFace dataset used for training generative models, sampled in ratios used for training. The bottom row shows generated data from models trained on the 80B-20W, 50B-50W, and 20B-80W datasets (left to right).}
\label{fig:real-and-gen-data}
\end{figure*}

\noindent\paragraph{Flickr Faces HQ (FFHQ)}is one of the most commonly-used datasets for the development of high fidelity GANs that generate faces, such as StyleGAN \cite{karras2019style} and StyleGANV2 \cite{karras2020analyzing}. It contains 70,000 high-quality 1024x1024 resolution images scraped from Flickr, which were automatically aligned and cropped \cite{karras2019stylebased}. We quantify the racial distribution of face images on a subset of FFHQ, as well as that of a StyleGANV2 model trained on FFHQ with and without truncation.

\noindent\paragraph{FairFace} consists of 108,501 images annotated by age, gender and race, which are cropped, aligned and $224 \times 224$ \cite{karkkainen2021fairface}. The representative faces in this dataset come from public images without public figures, minimizing selection bias.  
\new{As described in section~\ref{subsec:race}, we use the FairFace race categorization choices as a starting point. }

We use this dataset to train a \new{perceived} race classifier and multiple StyleGAN2-ADA models. From the FairFace dataset, we create three overlapping subsets of 12K images, each with images randomly sampled in the following ratios:
80\% Black/20\% white (80B-20W), 50\% Black/50\% white (50B-50W), and 20\% Black/80\% white (20B-80W). Six different $128\times128$ resolution StyleGAN2-ADA models \cite{karras2020training} were trained using an ``other races" dataset, an all Black dataset, an all white dataset, and each of the three FairFace datasets described above (i.e., 80B-20W, 50B-50W, and 20B-80W). All StyleGAN2-ADA models were trained with $2$ GPUs, batch size $64$, and an R1-gamma parameter value $0.0128$, ensuring high fidelity images for $25000$ kimg. Example generated images trained with the FairFace datasets can be seen in the bottom row of Fig.~\ref{fig:real-and-gen-data}.

\begin{figure}[t]
  \centering
\includegraphics[width=\textwidth]{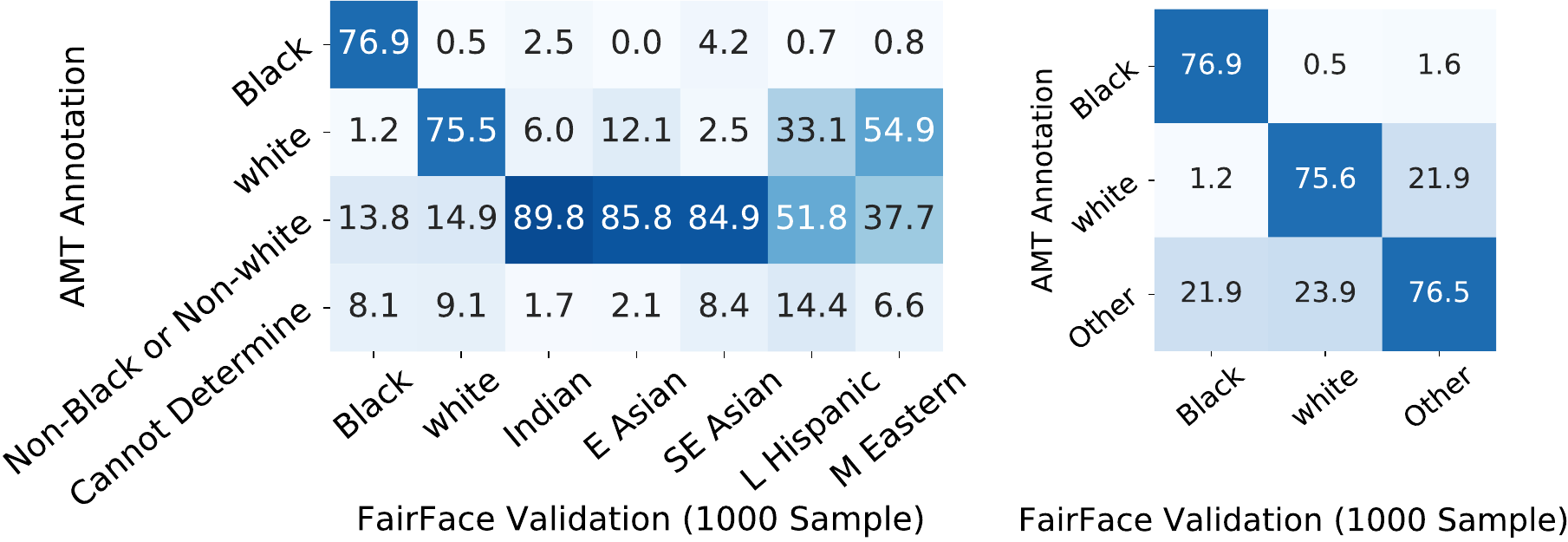}
\caption{{\bf Consistency of AMT annotation labels vs. Full FairFace labels.} (Left) A confusion matrix of AMT annotations and the 7 FairFace labels on 1000 random samples of the FairFace validation set. (Right) A condensed version of the confusion matrix. All numbers are shown in percentages. Overall, there is $76\%$ agreement between the FairFace labels and our collected annotations.}
    \label{fig:HL_FFace_performance}
\end{figure}

\subsection{Amazon Mechanical Turk Annotation}
\label{sec:amt}

Amazon Mechanical Turk (AMT) workers labeled tasks associated with the following three questions: 
\begin{enumerate}
    \item \textbf{Race Classification:} What is the race of the person in the image? [Choices: Black, white, Cannot Determine, Non-Black or Non-white.]
    \item \textbf{Real/Fake Classification:} Is this image real or fake? 
    \item \textbf{Image Quality Ranking:} Which image is more likely to be a fake image? 
\end{enumerate}

In our tasks, 1000 randomly sampled real images from FairFace and FFHQ, respectively, and 1000 images from each dataset of generated images were labeled. More details are in the supplementary material.

\section{Experiments and Results}

In this section, we first establish the reliability of our AMT annotation process and how we condense and use race labels, and then use the AMT annotations to compute the racial composition of a subset of FFHQ. We then assess the relationship between the racial distributions of the training and generated data, and evaluate the impact of truncation on the racially imbalance in the data. Finally, we assess the relationship between the training data racial distribution and the \new{perceived} image quality of the GAN-generated images.

\subsection{The Racial Distribution of Training and GAN-Generated Data}

\subsubsection{Annotation Consistency Analysis}
\label{sec:annotation_consistency}
We use annotations to measure the racial distribution of the training and generated images. We first assess the performance and reliability of our procedure by collecting annotations from a random sample of 1000 images from the FairFace validation set. The confusion matrix in Fig.~\ref{fig:HL_FFace_performance} shows the difference in labels on the FairFace validation set and the annotations collected using our AMT protocol, demonstrating the inherent limitations in attempting to establish racial categorizations on visual perception alone. However, we find that the ambiguity in visual discernment is lowest between \new{images perceived as Black and white, making these two classes suitable for analysis of racial bias.} 
Limiting the observed racial categories leads to more consistent labeling, allowing for a thorough examination of the impacts between two race class labels from the data.

\begin{figure}[t]
	\begin{center}
	\begin{tabular}{cc}
	 \includegraphics[height=0.2\textheight]{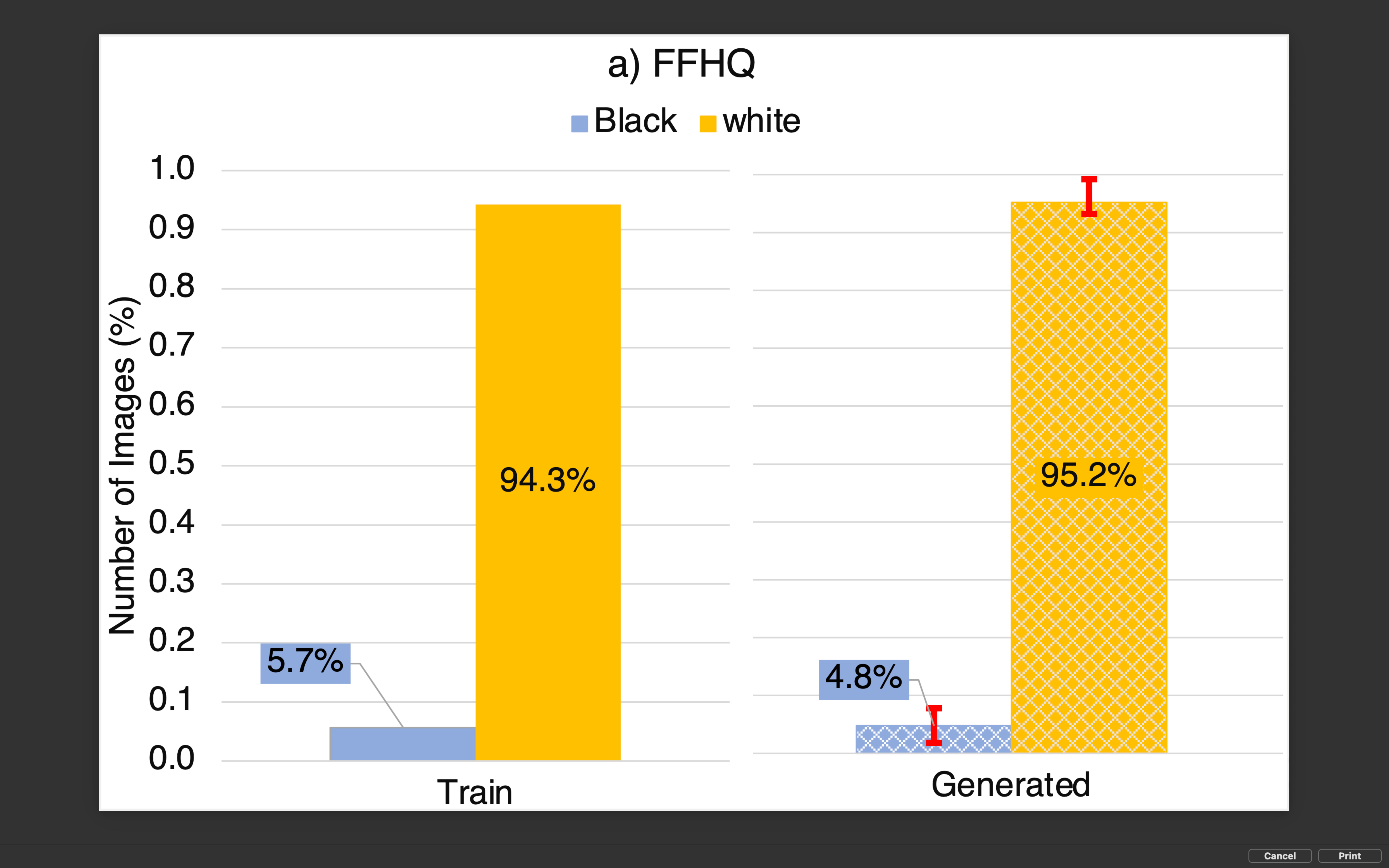} \hfill
      &
	 \includegraphics[height=0.2\textheight]{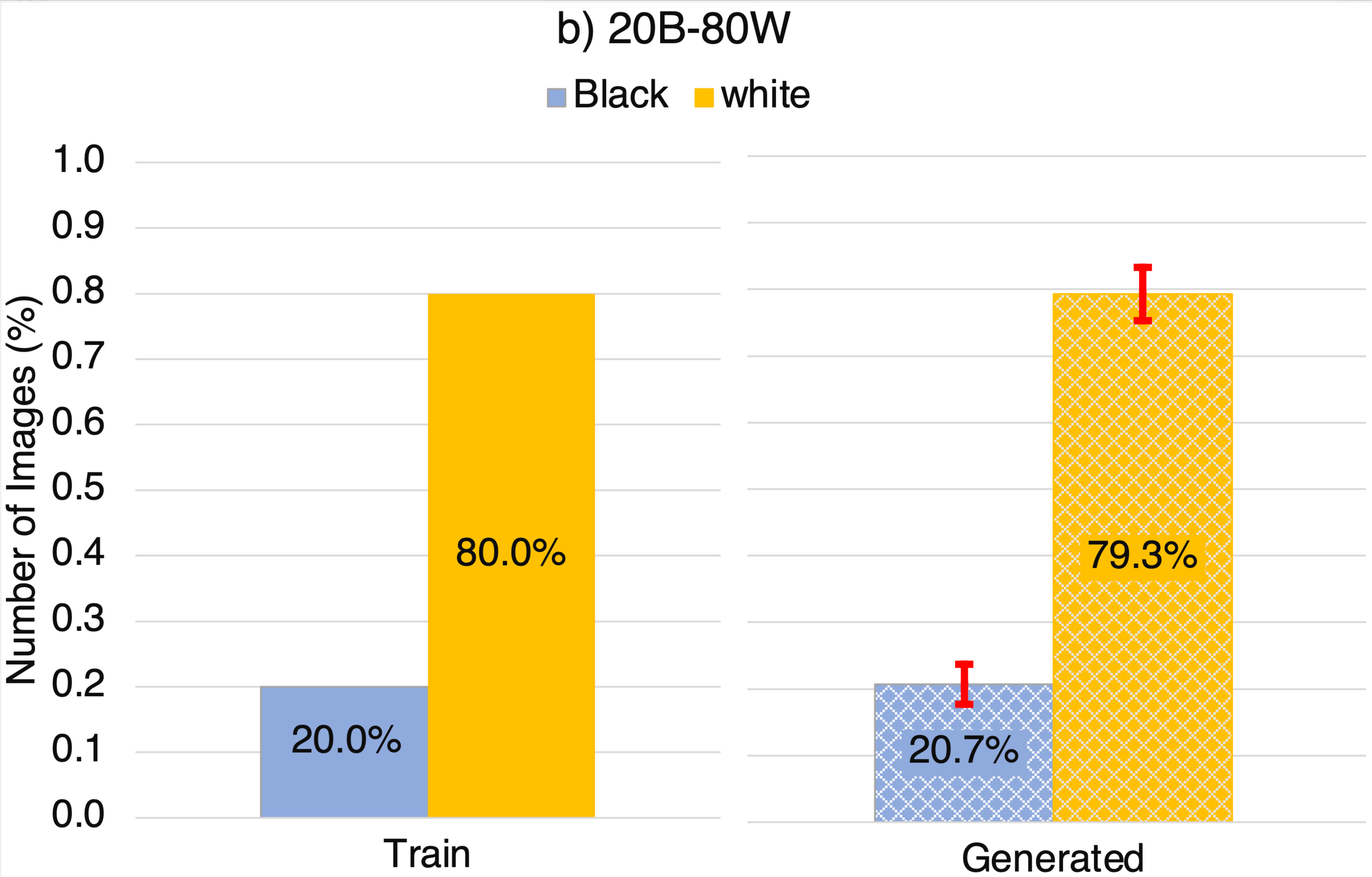}
	\\
	 \includegraphics[height=0.2\textheight]{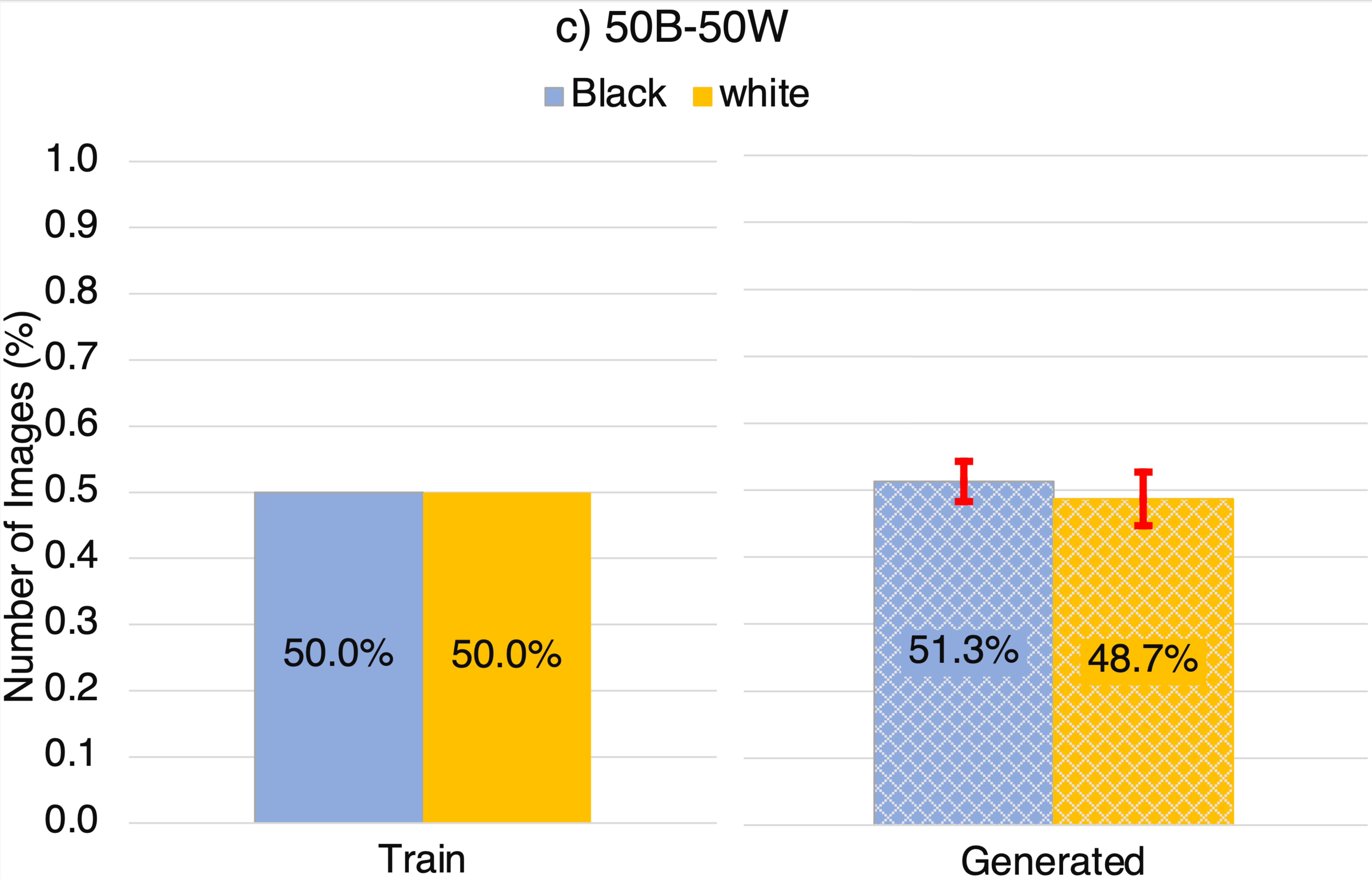} \hfill
	 &
	 \includegraphics[height=0.2\textheight]{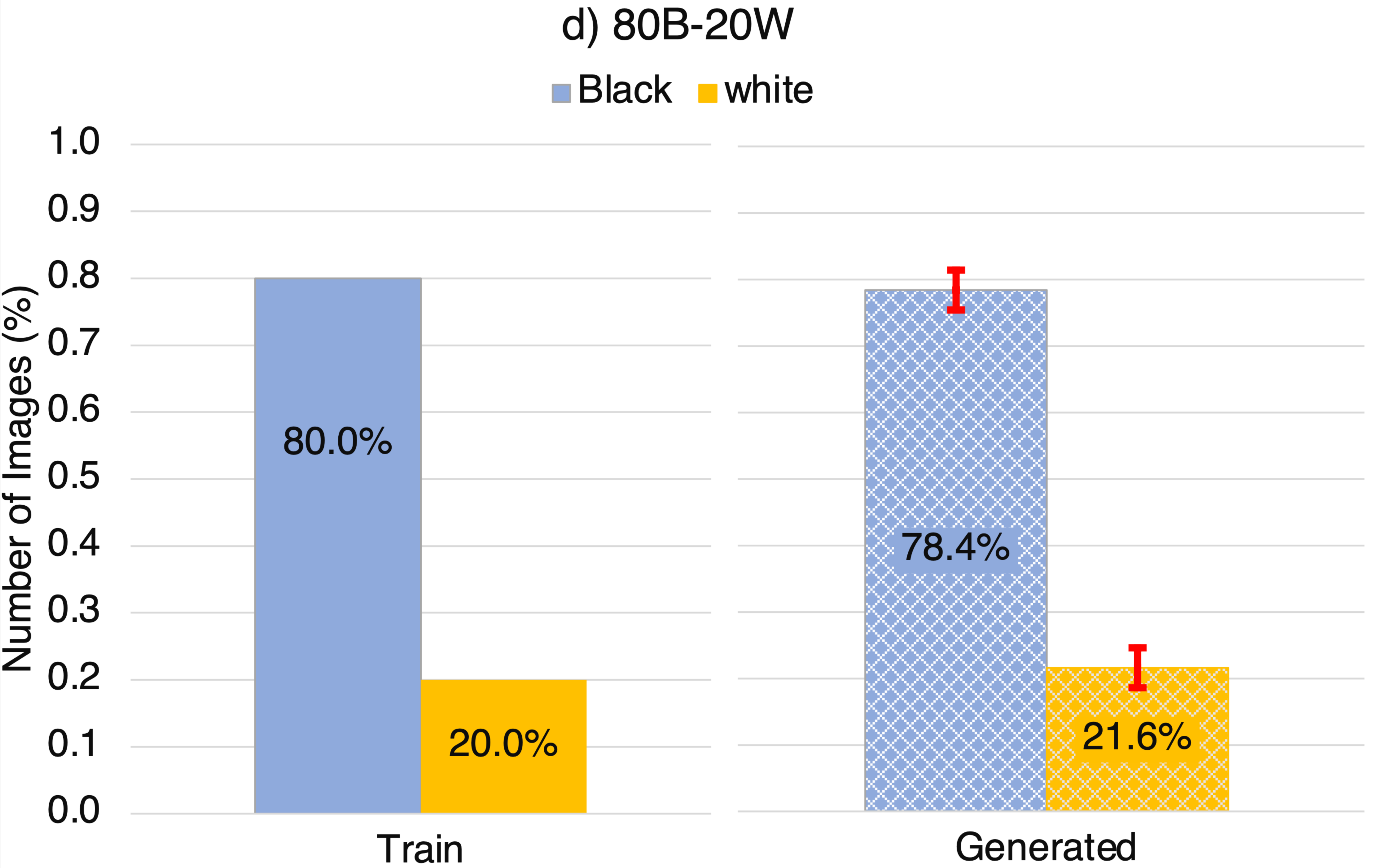}
	 \end{tabular}
    \end{center}
    \caption{{\bf Racial distribution of training and GAN-generated data.} Distributions for (a) FFHQ, (b) 20B-80W, (c) 50B-50W and (d) 80B-20W. The red bars represent the 95\% confidence interval for the expected distribution of the generated data. All class labels aside from Black and white are excluded. All of the generative models preserve the distribution of the training data.}
    \label{fig:fface_val_train_gen}
\end{figure}

\begin{figure}[t]
  \centering
	\begin{tabular}{c}
	 \includegraphics[height=0.4\textheight]{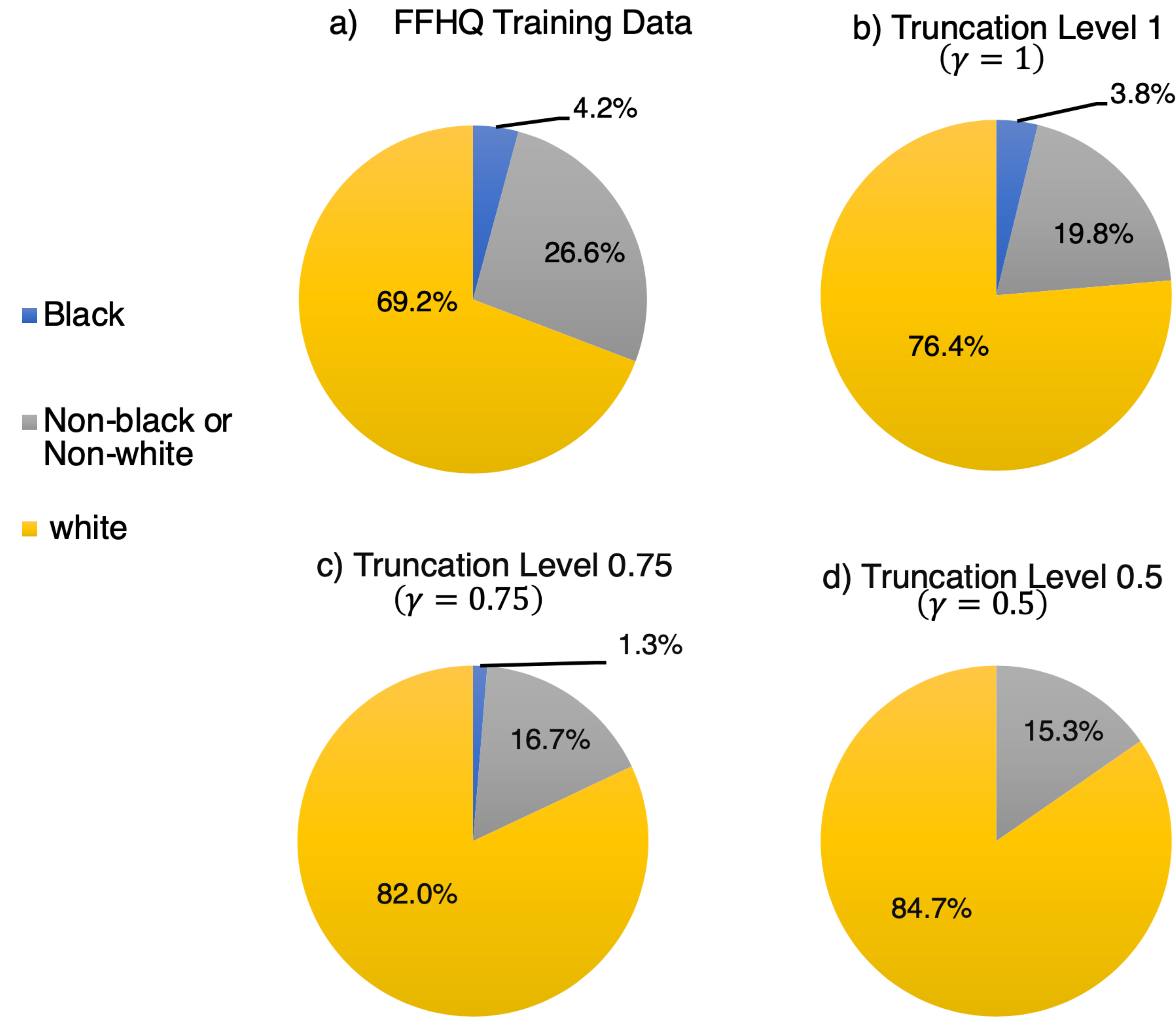}
	\hfill
	 
	 \end{tabular}
    \caption{{\bf The racial distribution of FFHQ with truncation}. The top row shows the FFHQ training data distribution (left) and the generated data distribution without truncation (right). The bottom row shows the generated data distribution with a truncation level of 0.75 (left) and an increased truncation level of 0.5 (right). While the model without truncation closely preserves the original training data distribution, as the level of truncation increases, the ratio of white to Non-white \new{class labels} increases.}
    \label{fig:ffhq_trunc_hl}
\end{figure}

\begin{figure}[h]
  \centering
	\begin{tabular}{ccc}
	  a) FairFace Validation & 
 b) AMT annotations \\
	  \includegraphics[height=0.2\textheight]{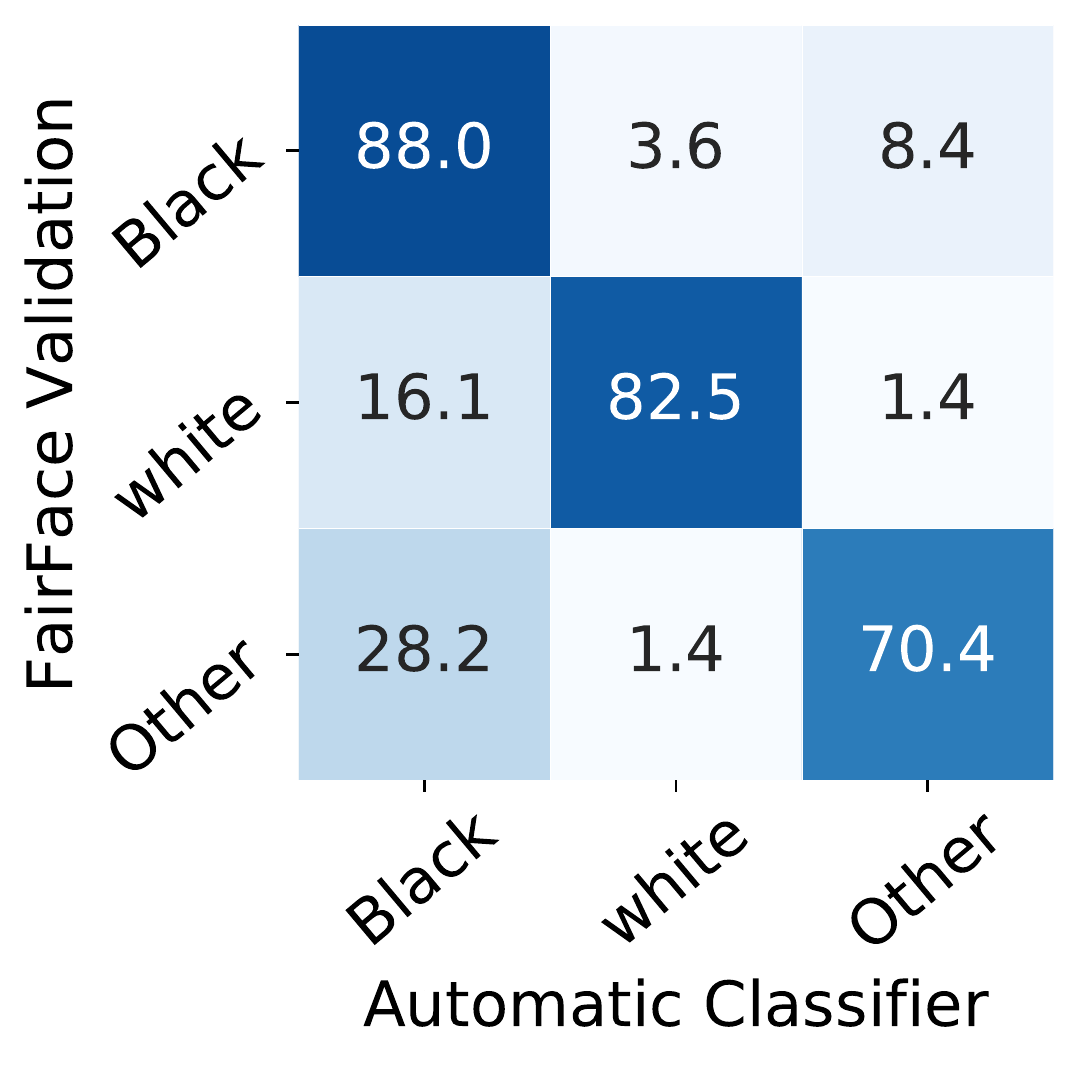} &
	  \includegraphics[height=0.2\textheight]{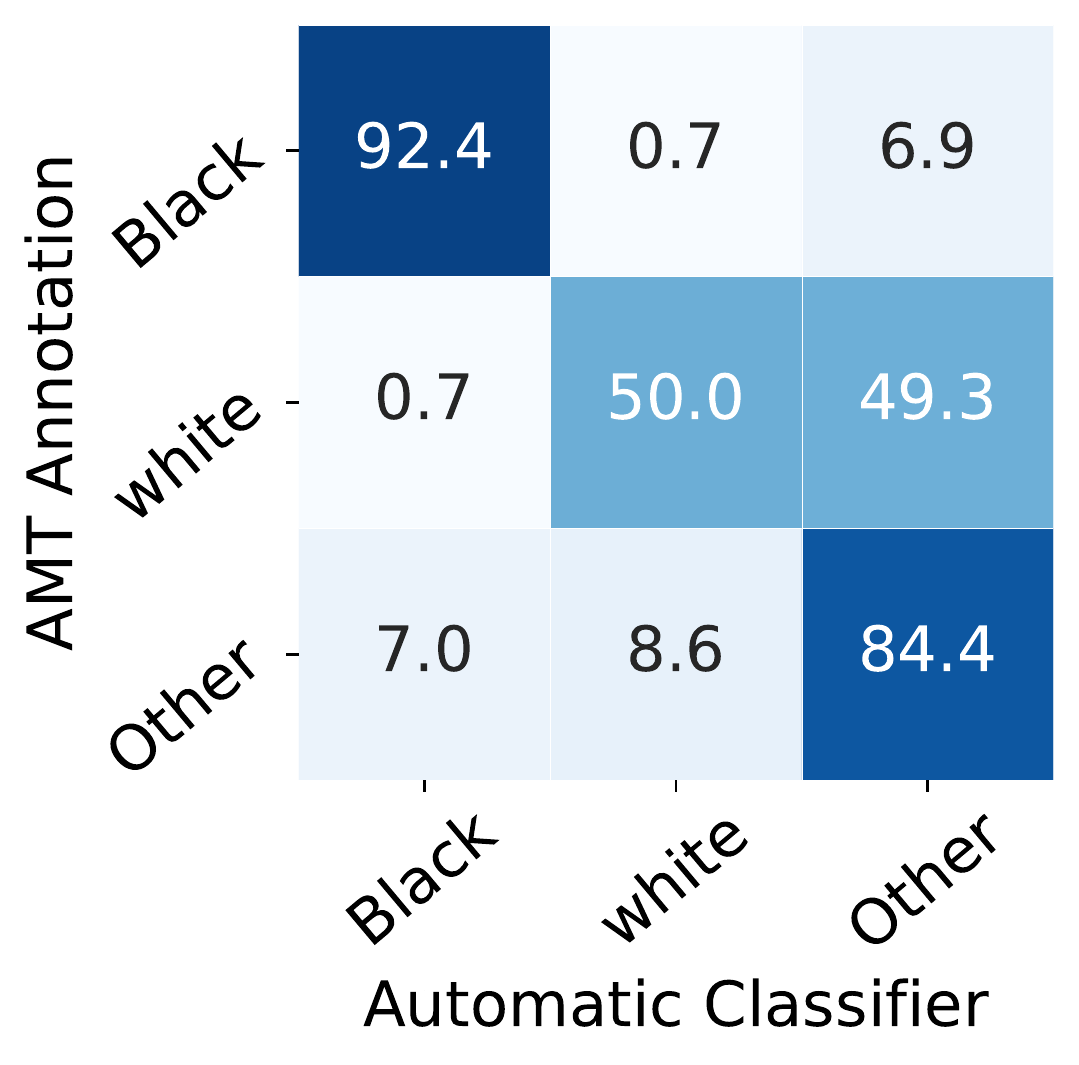} \\
    
    \end{tabular}
    \caption{{\bf Automatic Classifier Performance.} Confusion matrices for the automatic classifier on the entire FairFace validation set (left) and the automatic classifier on $1000$ images with collected annotations (right). On our AMT annotations, the classifier confuses \new{images labeled as Black and white} at a level comparable to that of our annotations and the FairFace labels (Fig.~\ref{fig:HL_FFace_performance}).}
    \label{fig:auto-clf-performance}
\end{figure}

\subsubsection{The Racial Distribution of FFHQ}
\label{sec:ffhq_dist}
 We analyze the racial distribution of FFHQ by selecting a random subset of $1000$ images and collecting images on AMT using our procedure for task one described in Section~\ref{sec:amt}. We find that FFHQ is composed of $69\%$ white, $4\%$ Black, and $27\%$ non-Black or non-white \new{facial images}. Compared to the global Black population, FFHQ is under-representative.

\subsubsection{Relationship between Training and GAN-Generated Data Distributions} On our first research question, regarding if an imbalanced dataset further exacerbates the generated dataset distribution, our experiments indicate that StyleGAN2-ADA's generated data distribution preserves the training data distribution. We compute the \new{perceived racial} distributions of FFHQ and FairFace (20B-80W, 50B-50W, and 80B-20W) training data and generated data, based on our AMT annotations. To explicitly showcase the ratio of Black and white race class labels in the training and generated data, we excluded the ``Non-Black or Non-white" and ``Cannot Determine" class labels. It can be seen in Fig.~\ref{fig:fface_val_train_gen} that the training and generated data distributions are statistically close - the red bars represent the $95\%$ Wald's confidence interval (CI) of each generated data distribution. 
The training data distributions all fall within the $95\%$ confidence interval of the observed sample means, and as such we conclude that the generators successfully preserve the training distributions and exhibits data distribution bias.
See the supplementary material for more information on CI calculations.

\subsubsection{Impact of Truncation on FFHQ Generated Data Distribution }

This section studies our second research question on the effect of truncation on racial imbalance. We follow the same protocol as above on truncation levels 1 (no truncation), 0.75, and 0.5, and find that that applying truncation when generating data exacerbates the racial imbalance in StyleGAN2-ADA.
Fig.~\ref{fig:ffhq_trunc_hl}(a) shows the AMT annotation distribution of the FFHQ training data, and \ref{fig:ffhq_trunc_hl}(b) shows the distribution of StyleGAN-2 ADA trained on FFHQ without truncation. 
As greater truncation levels are applied, the generated data becomes increasingly racially imbalanced. The percentage of \new{images of Black people} in the generated data distribution in Fig.~\ref{fig:ffhq_trunc_hl} drops from 4\% to 0\% at a truncation level of $0.5$. We observe an inverse effect for \new{images of white people}, where more truncation increases the percentage of the white class labeled images in the generated data distribution.

\subsubsection{Automatic Race Classifier} In order to conduct a more fine-grained study on the effect of  truncation level, we scale the AMT annotation process by using an automatic race classifier \new{to classify perceived race}. A ResNet-18 model \cite{he2016deep} was used to carry out three-way classification on face images. The model was trained on the FairFace training split augmented with equal quantities of generated images from StyleGAN2-ADA models trained on all-Black, all-white, and all-``other"-races datasets. Confusion matrices showing the performance of the classifier on the FairFace validation set and our collected annotations are shown in Fig.~\ref{fig:auto-clf-performance}. While the automatic classifier performance is not perfect, at $84\%$ accuracy, it suffices as a reasonable proxy for AMT annotations, given that the confusion between the classifier labels and our collected annotations on \new{images labeled as Black and white} is similar to the confusion between our collected annotations and the FairFace labels seen in Fig.~\ref{fig:HL_FFace_performance}.

\subsubsection{Evaluation of Truncation} 
\begin{figure}[t]

\centering
\begin{tabular}{cc}
	\includegraphics[height=0.2\textheight]{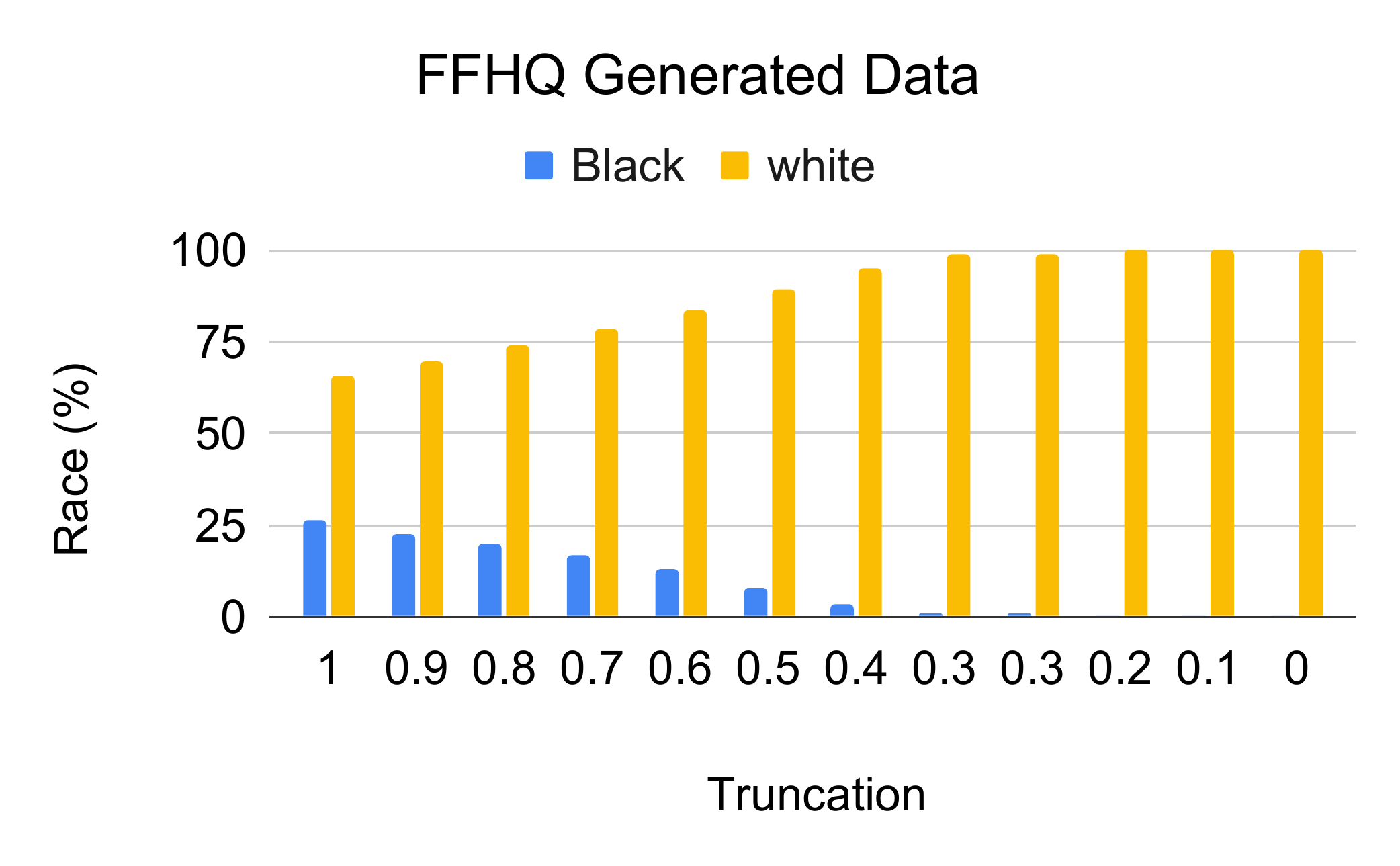} & 
	  \includegraphics[height=0.2\textheight]{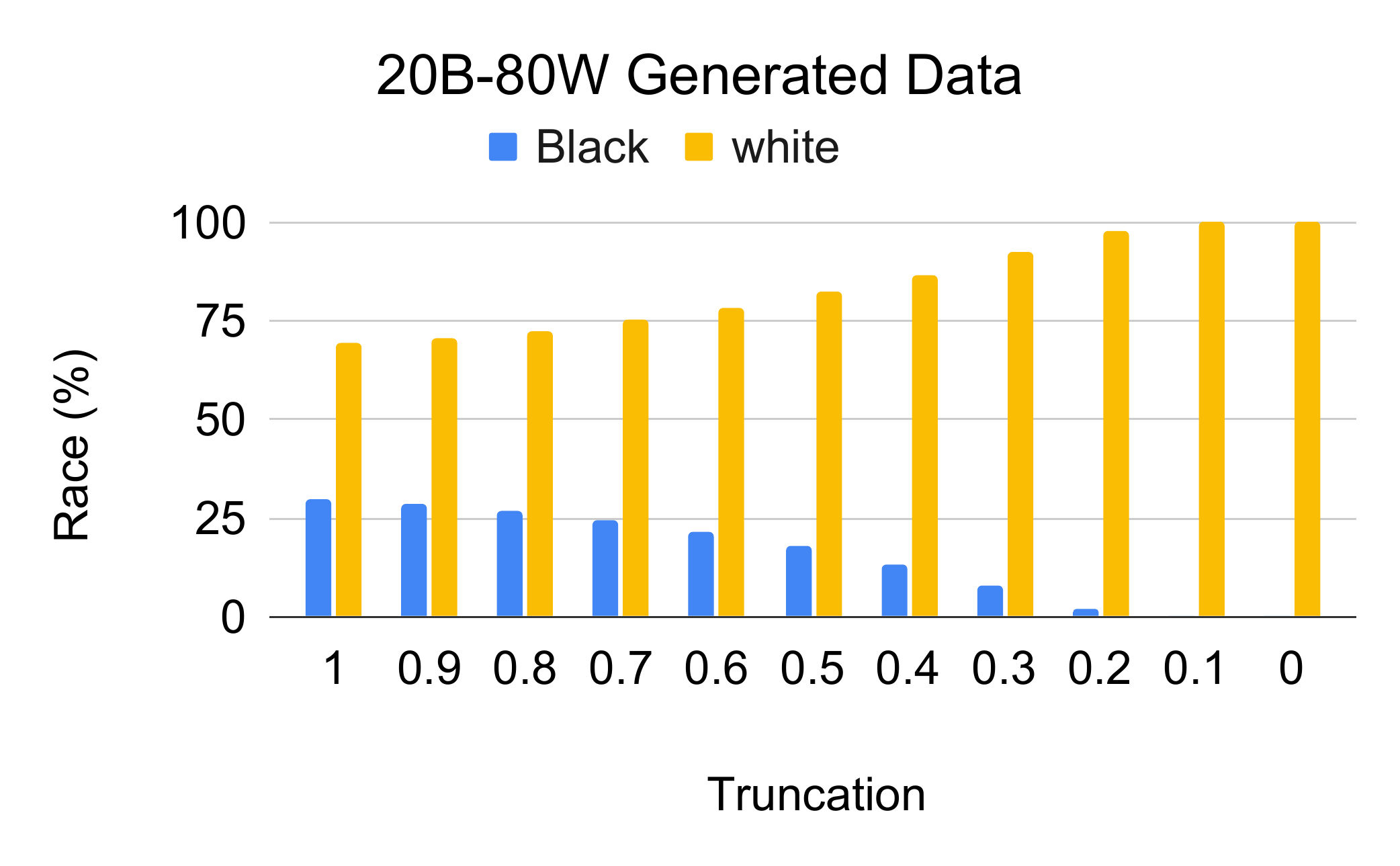} \\ 
	  \includegraphics[height=0.2\textheight]{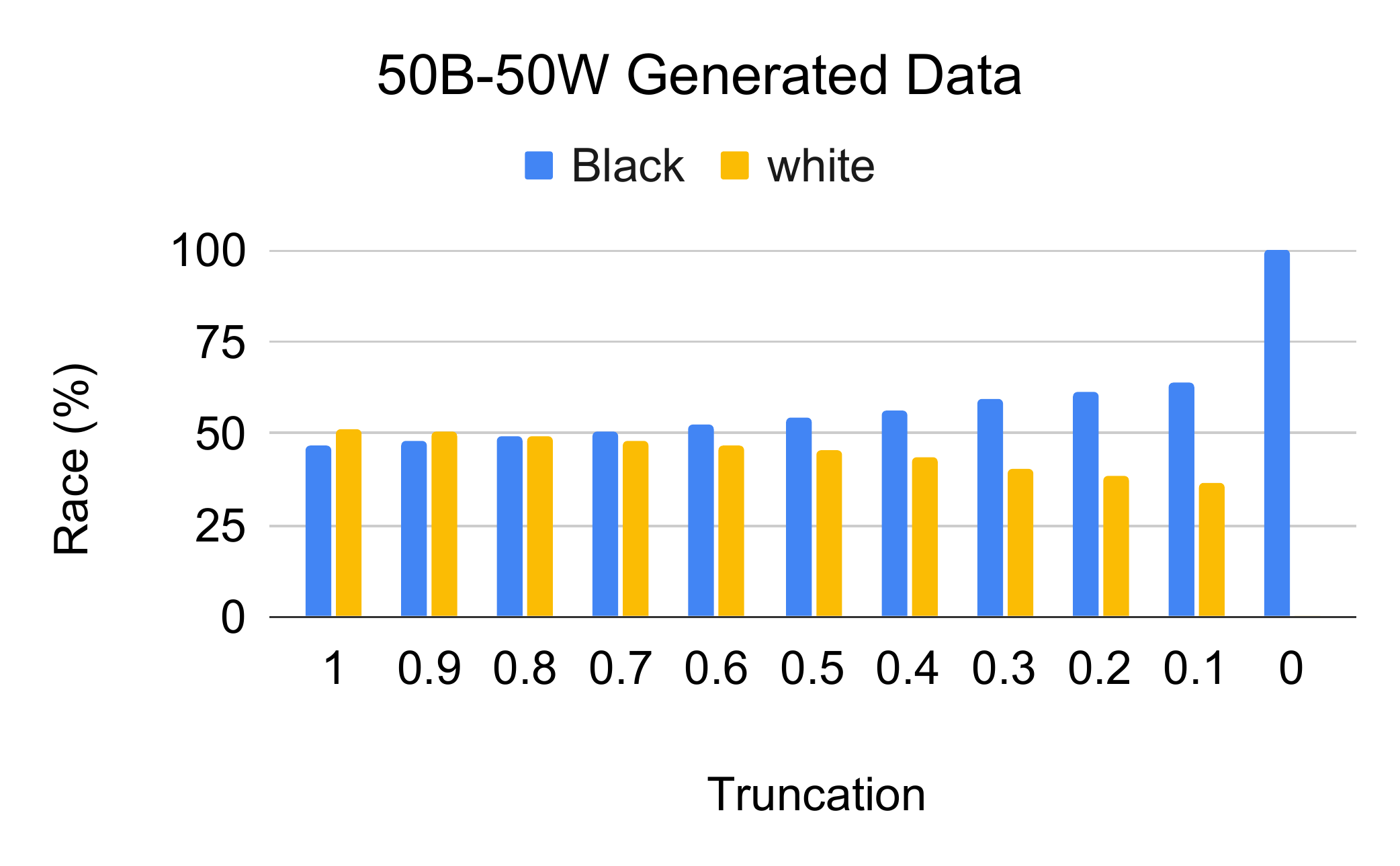} &
    \includegraphics[height=0.2\textheight]{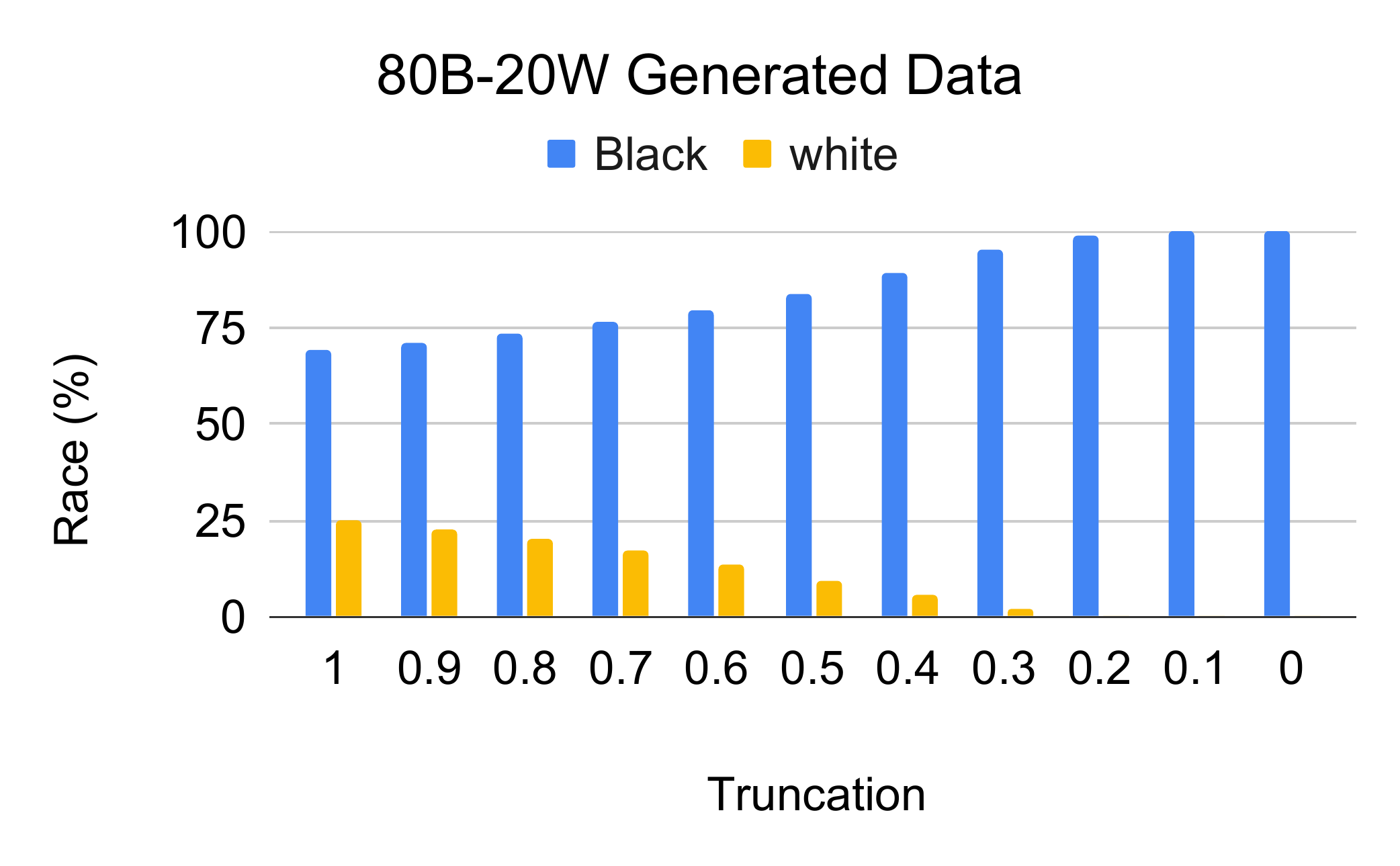} 
    \end{tabular}
    \caption{{\bf Automatic evaluation of truncation.} Automatically evaluated results of StyleGAN models trained on various datasets, with $110,000$ images total generated at levels of truncation from $\gamma = 1.0$ (no truncation) to $\gamma=0.0$ (full truncation). The y-axis represents the racial breakdown of the dataset, which becomes more polarized as truncation increases.} 
    \label{fig:FFHQ-trunc-auto}
\end{figure}

We evaluate levels of truncation and observe the following trend across all models: as the level of truncation increases, racial diversity decreases, converging to the average face of the dataset. Images were generated from StyleGAN2 trained on FFHQ and the 80B-20W, 50B-50W, and 80B-20W FairFace-trained generators at truncation levels ranging from $\gamma=0$ to $1$ at intervals of $0.1$. The \new{perceived} race labels were automatically classified for 10K generated images at each truncation level, for a total of 110K images, with results in Fig.~\ref{fig:FFHQ-trunc-auto}. We observe that truncation in a dataset with predominantly \new{images of white people}, such as FFHQ, increases the frequency of generating \new{images classified as white}. Similarly, when the majority of images in a dataset \new{are  of Black people}, as in the 80B-20W dataset, the truncated generated data distribution \new{has predominantly images classified as Black}. Examples of generated face images at different levels of truncation can be seen in the supplementary material.

\subsection{GAN Quality}

From the AMT Image Quality Ranking task, we perform pairwise comparisons of the generators' respective data distributions against each other. We determine that when trained with FairFace splits, images from the generator \new{trained on more images of white people} are always preferred at a greater proportion, and on FFHQ with truncation, images with more truncation are seen as higher quality. 

This finding comes from counting \new{the number of times images from one generator are preferred over images from another generator} 
Results of this comparison can be seen in Table~\ref{tab:generator-comparisons-ff} for FairFace and Table~\ref{tab:generator-comparisons-ffhq} for FFHQ. Error bars are computed using the Wald's method for a $95\%$ confidence interval. For FairFace, we find that the generator trained on a higher percentage of \new{images of white people} tends to be preferred in more comparisons by a narrow margin that often surpasses error bounds. For FFHQ with truncation, an increased truncation level always leads to a generator being more preferred, indicating that truncation increases perceptual sample quality.

\paragraph{Correlation with FID.} Fréchet inception distance (FID) \cite{fidscore} is a common GAN metric that measures image quality by comparing distributions of features extracted with the Inception network \cite{szegedy2015going} between training and generated images. Higher quality models receive a lower FID score.
The FIDs of our FairFace-trained generators are $5.60$, $5.63$, and $5.68$ for the 20B-80W, 50B-50W, and 80B-20W models respectively, not revealing a clear difference in perceived visual quality based on this automatic metric.

\begin{table}[t]
    \centering
   \caption{{\bf Pairwise image quality comparison of FairFace generators.} $9000$ comparisons were conducted between each pair of generators, resulting in a total of $27000$ comparisons. We report the percentage of images that are preferred from the left generator over the right, with the accompanying $95\%$ Wald's CI. Generators trained on datasets with a greater number of images of white people tend to be perceived as having better image quality.}
    \setlength{\tabcolsep}{12pt}
        \begin{tabular}{ccc}
            \toprule
            Generator A & Generator B & Percentage Gen. A Preferred \\

            \midrule
            20B-80W & 80B-20W & 53.9 $\pm$ 1.02     \\
            20B-80W & 50B-50W & 52.0 $\pm$ 1.03     \\
            50B-50W  & 80B-20W & 51.0 $\pm$ 1.03     \\
            \bottomrule
        \end{tabular}
    \label{tab:generator-comparisons-ff}
\end{table}

\begin{table}[t]
    \centering
     \caption{{\bf Pairwise image quality comparison of FFHQ at different truncation levels.} The percentage of images that are preferred when generated with the left truncation over the right, with a Wald's $95\%$ CI. A truncation level of $\gamma=1$ corresponds to no truncation, and $\gamma=0.5$ corresponds to the most truncation. Images generated with more truncation are perceived as being of higher quality.}
    \setlength{\tabcolsep}{12pt}
        \begin{tabular}{ccc}
            \toprule
             Truncation A &  Truncation B & Percentage Trunc. A Preferred\\
            \midrule
0.50  & 1.00    & 58.7 $\pm$ 1.02 \\
0.75 & 1.00    & 55.4 $\pm$ 1.02 \\
0.50  & 0.75 & 52.5 $\pm$ 1.03  \\
            \bottomrule
        \end{tabular}
    \label{tab:generator-comparisons-ffhq}
\end{table}

\subsection{\new{Perceived Visual Image} Quality and Race}
\begin{table}[t]
    \centering

            \caption{{\bf Top $K$ image composition per-race.} Given a ranking of images labeled as Black and white across all data splits, we break down the data split that each image came from. The highest quality images ($K=10, 25, 50$) are more likely to come from a data split where they are over-represented or represented in parity}
        \begin{tabular}{cccc}
       \multicolumn{2}{c}{} & \multicolumn{1}{c}{\textbf{white}} & 
            \multicolumn{1}{c}{} \\ 
            \toprule
            $K$ & 80B-20W &  50B-50W & 20B-80W \\

            \midrule
            10  & 0.00 & 0.29 & \textbf{0.71} \\
            25  & 0.00 & 0.38 & \textbf{0.62} \\
            50  & 0.22 & 0.36 & \textbf{0.42} \\
            100 & 0.26 & \textbf{0.41} & 0.33 \\
            500 & 0.33 & \textbf{0.36} & 0.31 \\ \bottomrule
        \end{tabular}
        \qquad
        \begin{tabular}{cccc}
        \multicolumn{2}{c}{} & \multicolumn{1}{c}{\textbf{Black}} & \multicolumn{1}{c}{} \\
            \toprule
            $K$ & 80B-20W &  50B-50W & 20B-80W \\
       
            \midrule
            10  & \textbf{0.49} & 0.23 & 0.28 \\
            25  & \textbf{0.57} & 0.19 & 0.24 \\
            50  & \textbf{0.45} & 0.37 & 0.17 \\
            100 & \textbf{0.37} & 0.32 & 0.31 \\
            500 & \textbf{0.35} & 0.33 & 0.32 \\
            \bottomrule
        \end{tabular}
    
    \label{tab:quality-race-breakdown}
\end{table}

To address the third research question, i.e., to determine if there is a relationship between \new{perceived} race and generated image quality, we examine the results of our binary real/fake classification task and our pairwise image quality ranking task. Our findings on the real or fake classification task do not yield a clear relationship between the training data distribution and generated image quality; please see the supplementary material for details. However, pairwise image quality comparisons provide a more fine-grained analysis. From a perceptual quality ranking obtained from pairwise comparisons, we find that the average \new{perceived visual} quality of generated images of a particular race  increases as the proportion of training images of that race increases. We also find that generated \new{images of white people} tend to be perceived as higher quality than \new{images of Black people}, regardless of the training distribution.

\begin{figure}[t]
    \centering
    \includegraphics[width=0.9\textwidth]{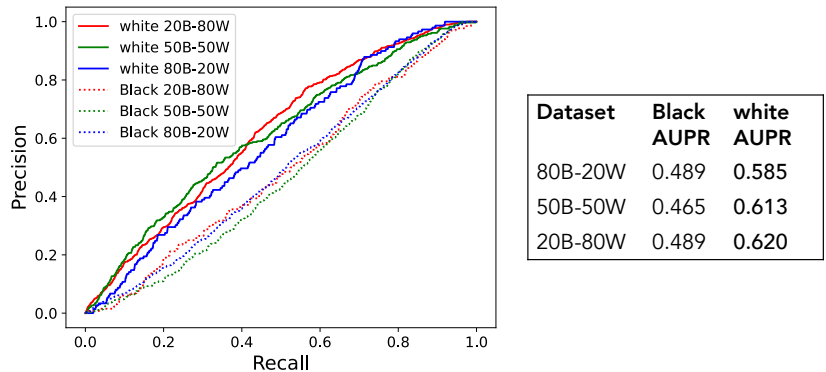}
    \caption{{\bf PR curves (left).} PR curves for each \new{race} within each generated dataset. For a given $K$ between $0$ and $3000$, precision, shown on the y-axis, is defined as the count of images of a particular \new{perceived} race and dataset in the top $K$ images, normalized by the total number of images of each race and dataset. Recall, shown on the x-axis, is defined as the number of images seen out of the total images ($K/N$, where $N = 3000$). {\bf Area under the PR curves (right).} A larger number indicates high image quality; \new{images labeled as white} are consistently perceived as higher quality than \new{those labeled as Black}, regardless of the generated dataset they come from.}
    \label{fig:global_ranking_counts}
\end{figure}

Using $3000$ FairFace 80B-20W, 50B-50W, 20B-80W dataset images, $54000$ pairwise comparisons were evaluated within and across the datasets. From these pairwise comparisons, we use the \texttt{choix} package's \cite{choix_git} implementation of the Bradley-Terry model \cite{caron2012efficient} to rank the $3000$ images in descending order of image quality. From this global ranking, we obtain a ranked ordering of all images labeled as Black and white. Table~\ref{tab:quality-race-breakdown} investigates the breakdown of the top $K$ images. In order to obtain weighted percentage scores, the raw counts for the top $K$ images (which can be seen in the supplementary material) of a particular race are normalized by the expected frequency of the images from the corresponding race found in each data split.
Then, the weighted numbers are divided by the sum of all scores for that race and value of $K$. The results indicate that the highest quality images of a particular race are more likely to come from a data split where the race class is over-represented or represented in parity. From the global ranking, a precision-recall curve for each race from each data split over the top $K$ images, and the area under the PR curves, are shown in Fig.~\ref{fig:global_ranking_counts}. Images \new{labeled as white} are overall ranked as higher quality than images \new{labeled as Black}. Furthermore, for white labels, being in the majority (i.e., from the 20B-80W split) yields better quality than in the minority (i.e., from the 80B-20W split).

These results raise the question of whether a predisposition towards white generated faces is a by-product of our learned generative models, or is a result of other parts of our data collection and evaluation process.
In order to gain insight on this question, we conducted $1700$ pairwise comparisons between real \new{face images from the FairFace data labeled as Black and white}, using the same AMT protocol as for generated data.
By removing generative models from this evaluation, we can determine whether external factors such as original real image quality or annotator bias may play a role in our observed results.
An evaluation procedure invariant to the \new{perceived} races of images should produce results where \new{real images perceived as white} are preferred over \new{real images perceived as Black} $50\%$ of the time. Instead, they were preferred $55.2\%$ of the time with a $95\%$ Wald's confidence interval of $55.2\% \pm 2.3\%$. This indicates that even though our system of evaluation is based on pairwise comparisons, a standard and well-regarded GAN quality metric \cite{borji2019pros}, it has a detectable bias towards selecting \new{images labeled as white over those labeled as Black}. 

The source of this propensity towards selecting \new{images perceived as white} is unclear. Captured images of \new{Black people} in the dataset could be of lower quality than that of \new{images of white people}, potentially because of camera or sensor bias. Due to prevalence, collecting high quality \new{images of white-appearing faces} might be easier. Another possibility is the ``other race effect" \cite{Phillips2011}, where annotators are biased toward their own race, however, the demographics of the annotators in our study are unknown. A future in-depth study of these factors \new{causing asymmetric algorithmic bias} should be a subject of future investigations.

\section{Discussion}

Through a systematic investigation into the role of racial composition in generative models, we find that state-of-the-art GANs such as StyleGAN2-ADA closely mirror the racial composition of their training data, \new{exhibiting data distribution bias}. Our study reveals that in FFHQ, the most prominent dataset of generative models of \new{facial images, Black people} are substantially underrepresented at $4\%$, as compared to the global population. Practitioners should be aware of this bias when using this dataset. We recommend that generative modeling practitioners have more awareness of the racial composition of their training dataset, particularly when the downstream application requires a well-balanced model. When the training data has known biases or imbalances, we recommend transparency through mechanisms such as model cards \cite{gebru2018modelcards}.

Downstream applications, even generative model demos, often employ the truncation trick to improve the visual fidelity of generated images. Our work shows that this qualitative improvement comes at the expense of exacerbating existing bias. Our studies show that using a well balanced dataset can mitigate this issue \new{of symmetric algorithmic bias}. We suggest researchers be transparent on their usage and level of truncation,  and encourage research for alternative algorithms to truncation. Interesting future directions include correlating FID to other quality metrics, performing this study on different GAN architectures and other generative models such as VAEs \cite{kingma2013auto,razavi2019generating} and diffusion models \cite{ho2020denoising}. \new{In particular, with diffusion models, it would be interesting to see if classifier-free guidance \cite{ho2022classifier} exhibits the same symmetric algorithmic bias as the truncation trick. Another interesting direction is to perform an intersectional study probing similar questions, by considering other attributes such as gender in addition to race}.

\subsubsection{Acknowledgements} 
We thank Hany Farid, Judy Hoffman, Aaron Hertzmann, Bryan Russell, and Deborah Raji for useful  discussions and feedback. This work was supported by the BAIR/BDD sponsors, ONR MURI N00014-21-1-2801, and NSF Graduate Fellowships. The study of annotator bias was performed under IRB Protocol ID 2022-04-15236.

\clearpage

\bibliographystyle{splncs04}
\bibliography{2581}
\end{document}


\pagestyle{headings}
\mainmatter
\def\ECCVSubNumber{2581}  

\title{Supplementary Material: Studying Bias in GANs \\
through the Lens of Race} 
\titlerunning{Supplementary Material: Studying Bias in GANs \\
through the Lens of Race} 
\authorrunning{V.H. Maluleke, N. Thakkar, et al.} 
\author{Vongani H. Maluleke*
\and
Neerja Thakkar*
\and
Tim Brooks 
\and 
Ethan Weber 
\and
Trevor Darrell
\and
Alexei A. Efros
\and
Angjoo Kanazawa
\and
Devin Guillory}

\institute{UC Berkeley}

\maketitle

In this supplementary material document, we first discuss the selected perceived race label categorization.
We then describe in detail the three Amazon Mechanical Turk tasks and implementation information. Followed by an analysis of the performance of both AMT annotations and the automatic classifier in evaluating perceived race. Visualizations of various truncation levels are shown next, and finally, we present more details and visualizations of quality ranking. 

\section{Race Labels Categorization}

\new{We start with the the FairFace dataset labels, and then collect annotations based on our own condensed categorization. 
The FairFace dataset started with the commonly accepted race categories from the U.S. Census Bureau---white, Black, Asian, Hawaiian and Pacific Islanders (HPI), Native Americans (NA), and Latino. They dropped the HPI and NA categories due to insufficient image examples, and expanded the Asian category into four distinct subgroups: Middle Eastern, East Asian, Southeast Asian, and Indian \cite{karkkainen2021fairface}. To reduce perceptual ambiguity (see main paper in section 4.1), we condense the race class labels from seven FairFace classes to three classes---Black, white, and Non-Black or Non-white---where Non-Black or Non-white comprises the Middle Eastern, East Asian, Southeast Asian, Latino Hispanic, and Indian as labeled by FairFace. We also relabel all the images we analyze using our own annotation protocol with three categories and a ``Cannot Determine" category.}

\section{Amazon Mechanical Turk (AMT) Details}
Amazon Mechanical Turk (AMT) was used to collect annotations for three label tasks, namely; (1) race classification, (2) quality classification, and (3) quality ranking. As mentioned in the paper, these tasks consist of the following questions:

\begin{enumerate}
    \item \textbf{Race Classification:} What is the race of the person in the image?
    \item \textbf{Real/Fake Classification:} Is this image real or fake?
    \item \textbf{Image Quality Ranking:} Which image is more likely to be a fake image?
\end{enumerate}

\mypar{Implementation details.} Our label tasks were deployed using a custom framework for deploying AMT tasks using our dynamically populated HTML/JavaScript template and the Python API Boto3\footnote{\url{https://boto3.amazonaws.com/v1/documentation/api/latest/index.html}}. Our code enables creating human intelligence tasks (HITs) that show images with a corresponding question, and then the annotator completes a forced-choice answer among a set of specified choices. For quality control, we use both \textbf{accuracy} and \textbf{consistency} checks. As an accuracy test, workers must get eight of these hidden questions correct. As a consistency test, we duplicate these ten test cases and scatter them throughout the HIT, and the worker must be consistent for eight of the repeated examples. Our hidden test cases are chosen to be adequately obvious such that diligent workers will successfully pass them. If less than eight are answered correctly, the worker's responses are discarded. Furthermore, we ensure that three unique workers answer each question. Each HIT starts with a consent form and a comprehensive description of the task with practice examples with accompanying answers and descriptions; this helps ensure annotators understand the tasks so they can pass our quality control checks.

Next, we go through each of the three tasks and provide qualitative examples for the questions being asked. In total, we asked a total of 50K questions in 1422 unique HITs, where each was labeled by 3 different workers.We did not collect the demographics or other information on AMT workers. Overall, 59 annotators participated in our tasks, and we paid on average \$8.71 per hour. Note that \textit{we will release our code used to conduct our experiments for the benefit of the computer vision community}.

\subsection{Race Classification}
The Race Classification AMT task asked the workers to identify the race of the person(s) in the image, by selecting one of the options described below:
\begin{enumerate}
    \item Black - This is an image of a Black person.
    \item white - This is an image of a white person.
    \item Non-Black or Non-white - I know the race of the person and the person is not Black or white.
    \item Cannot Determine - I cannot tell the race of the person.
\end{enumerate}

\begin{figure}[h]
	\begin{center}
	\begin{tabular}{cc}
	\subcaptionbox{Black}
	 	 {\includegraphics[height=0.25\textheight]{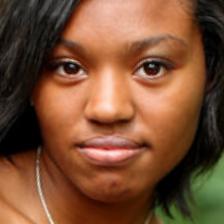}}	& 
		\subcaptionbox{White} {\includegraphics[height=0.25\textheight]{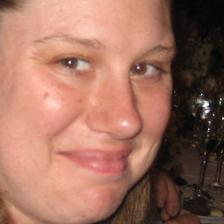}} \\ \\
	 	\subcaptionbox{Non-Black or Non-white} {\includegraphics[height=0.25\textheight]{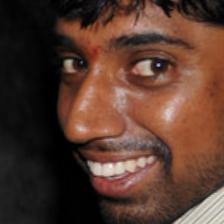}} & 
	\subcaptionbox{Cannot Determine} {\includegraphics[height=0.25\textheight]{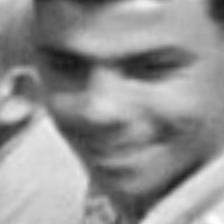}}
    \end{tabular}
    \end{center}
    \caption{Sample of the race classification examples with corresponding race class labels given to AMT workers before they started Task 1.}
    \label{fig:task1_examples}
\end{figure}

\begin{figure}[h]
	\begin{center}
	\begin{tabular}{c}
	 \includegraphics[height=0.25\textheight]{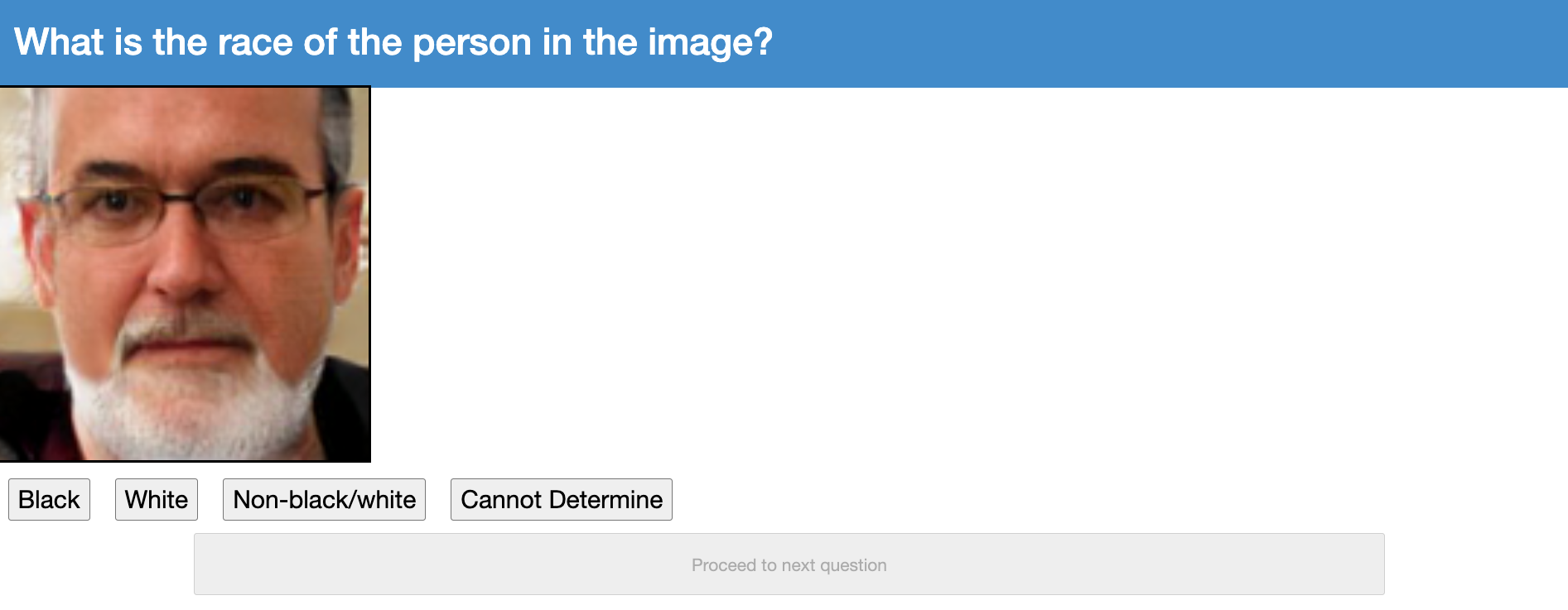} 
    \end{tabular}
    \end{center}
    \caption{Race Classification AMT Interface Example.}
    \label{fig:task1}
\end{figure}

\paragraph{} The AMT workers were given examples of images and corresponding race class labels- \ref{fig:task1_examples}, as well as a demo of the interface before they could start the task. Fig.~\ref{fig:task1} is an example of the deployed Race Classification AMT interface.

\subsection{Quality Classification}
The Quality Classification AMT task asked workers to identify real photographs and fake images by selecting one of the options described below:
\begin{enumerate}
\item Real Photograph - This is a photograph of a real person taken using a camera.
\item Fake/Manipulated Image - This is a computer-generated image of a person(s) who do not exist.
\end{enumerate}

\begin{figure}[h]
	\begin{center}
	\begin{tabular}{c}
	 \includegraphics[height=0.25\textheight]{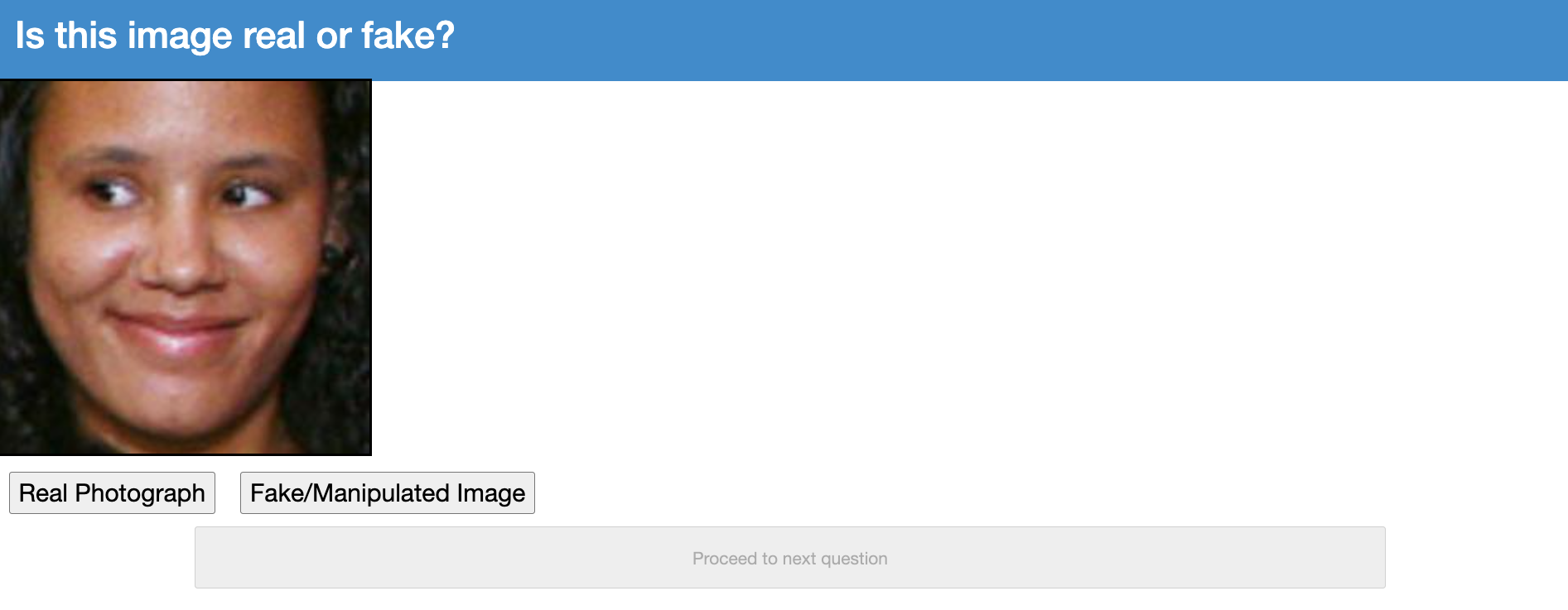} 
    \end{tabular}
    \end{center}
    \caption{Quality Classification AMT Interface Example.}
    \label{fig:task2}
\end{figure}

To further assist the workers in understanding the difference between the two options, the definition of each of the two options was provided to workers as ``Real photographs are images of real person(s) captured using a camera" and ``Fake/Manipulated images are computer generated images of a person(s) who do not exist". Fig.~\ref{fig:task2} is an example of the Quality Classification AMT interface.

\subsubsection{Results from Real/Fake Classification}
\begin{figure}[h]
	\begin{center}
	\begin{tabular}{ccc}
	 \includegraphics[height=0.25\textheight]{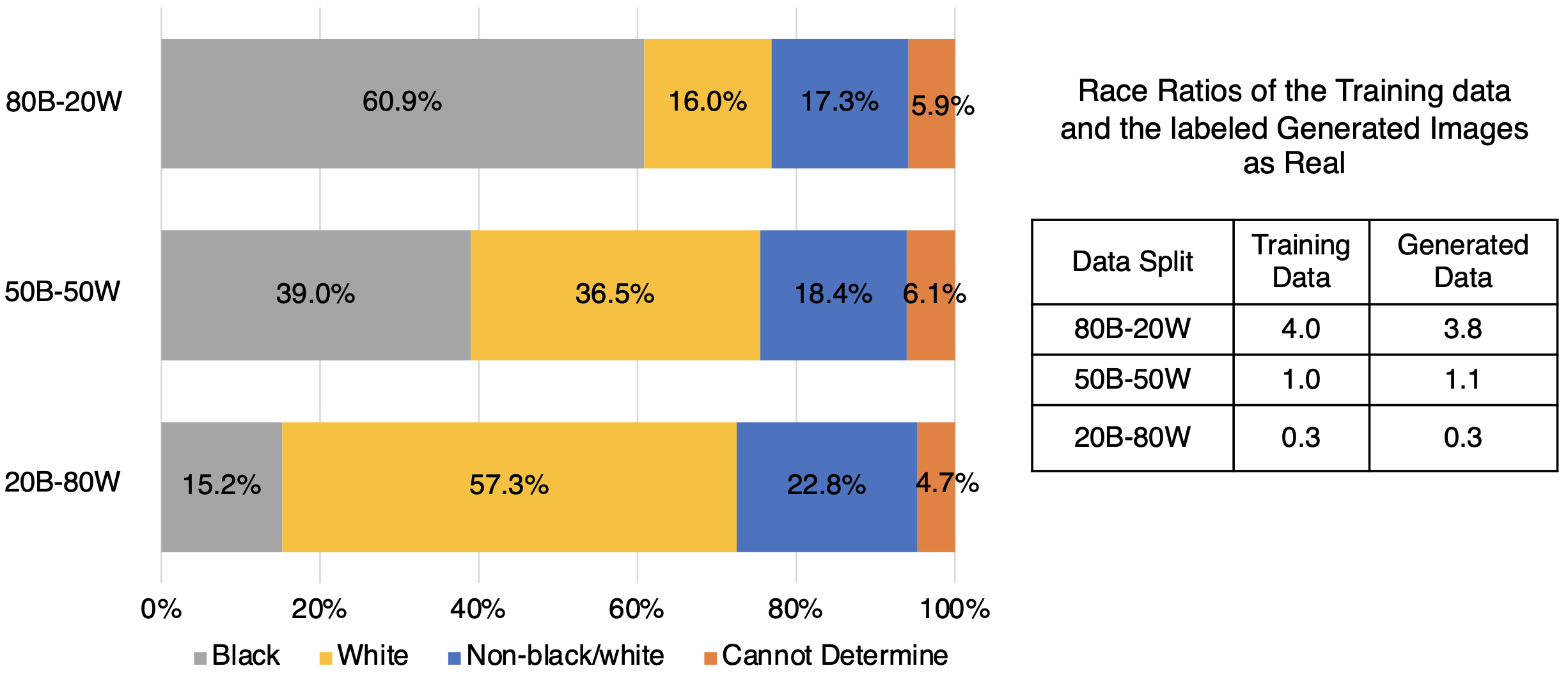} 
    \end{tabular}
    \end{center}
    \caption{Distribution of the Generated Images that were labeled as ``real", or high quality, images by annotators in the three FairFace data splits (20B-80W, 50B-50W and 80B-20W) with a corresponding table showing the race ratio (Black/white) of the different data splits. This shows that the race ratio of the training data is relatively the same as the race ratio of the generated images that were labeled Real.
     }
    \label{fig:real_fake}
\end{figure}

Using the Real/Fake indicator as a proxy for image quality, we are unable to determine any significant distinctions in generated image quality with respect to race. 
We use the label ``fake" as a proxy for low quality and ``real" to represent high quality. 
This image quality proxy measured was collected by using AMT annotators to determine if an image was real or fake. 
Fig.~\ref{fig:real_fake} shows the racial distribution of the generated images that were classified as real/high quality for the three data splits. 

In Fig.~\ref{fig:real_fake} we observe that the race class ratio of the training images has the same race class ratio as the training data. 
The race ratio of the 20B-80W, 50B-50W, and 80B-20W training data, respectively, has a race ratio of 0.25, 1, and 4, and the corresponding generated data race ratios of the high quality labeled images are 0.26, 1.1, and 3.8.

\subsection{Quality Ranking}
The Quality Ranking AMT task asked the workers to identify the Fake image between two images by selecting one of the options described below:
\begin{enumerate}
\item Image A - Image A is more likely to be a Fake image.
\item Image B - Image B is more likely to be a Fake image.
\end{enumerate}

\begin{figure}[h]
	\begin{center}
	\begin{tabular}{c}
	 \includegraphics[height=0.25\textheight]{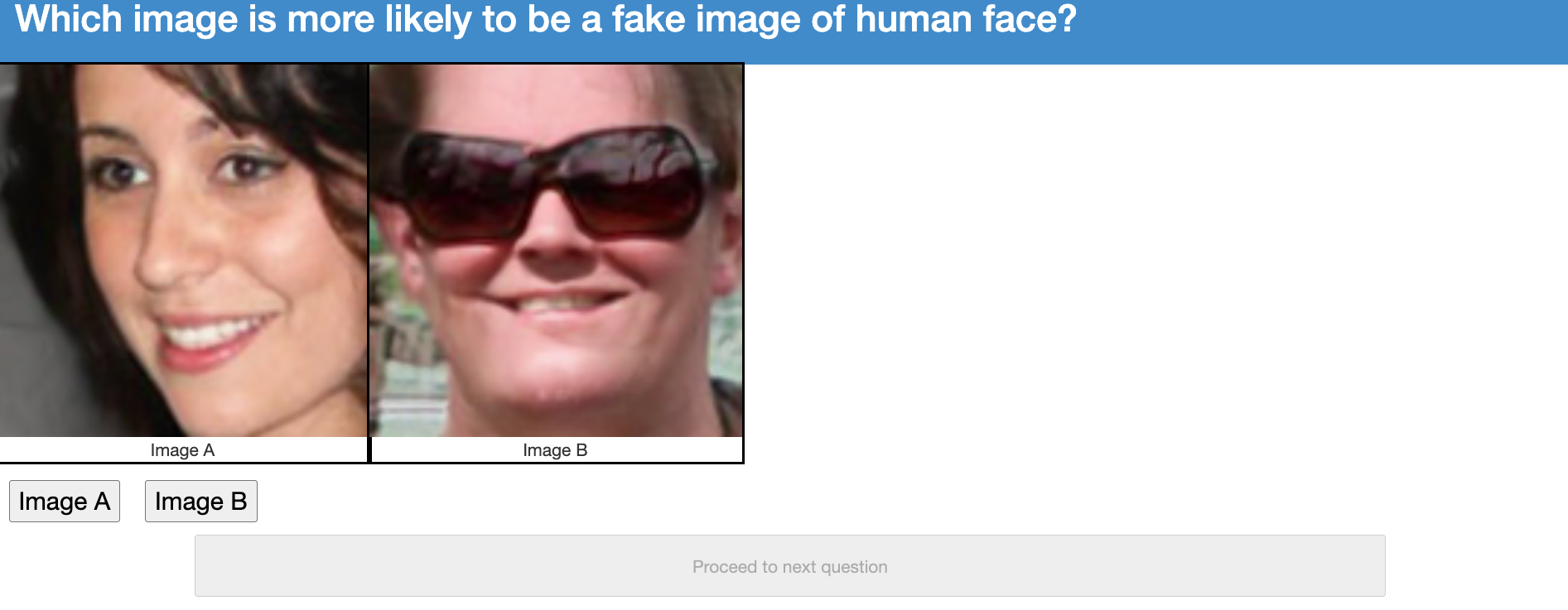} 
    \end{tabular}
    \end{center}
    \caption{Quality Ranking AMT Interface Example.}
    \label{fig:task3}
\end{figure}

To further assist the workers in understanding what Fake images are, the same definitions from above were given to the workers at the start of the task. Fig.~\ref{fig:task3} is an example of the Quality Ranking AMT interface.

\section{Race Classification Performance Analysis}
In this section we expand on the performance of the AMT annotation compared to the automatic perceived race classifier.

Our perceived race classifier obtained an accuracy of $84\%$, treating our FairFace labels as ground truth, on the same $1000$ images used for human annotation. This gave us more confidence on the performance and validity of the race classifier and its role as a proxy for AMT annotation when conducting our experiments. 

The classifier performs better on Black and Other class labels compared to the white class label. The classifier tends to classify white faces as Other which is also slightly observed in the AMT annotations.

Comparing the performance of the human annotation and the perceived race classifier we see that they are both aligned in terms of classifying the different race class labels, and therefore the automatic classifier can be used as a proxy for human annotation. Overall, the classifier outperforms the human annotations. We hypothesize that this could be due to subjective bias present in human annotation, or to the subjective nature of perceived race classification.

\section{Experimental Results}

\subsection{Relationship between Training and Generated Data Distributions} 
\begin{figure}[h]
\centering
\includegraphics[height=0.49\textheight,width = 1\textwidth]{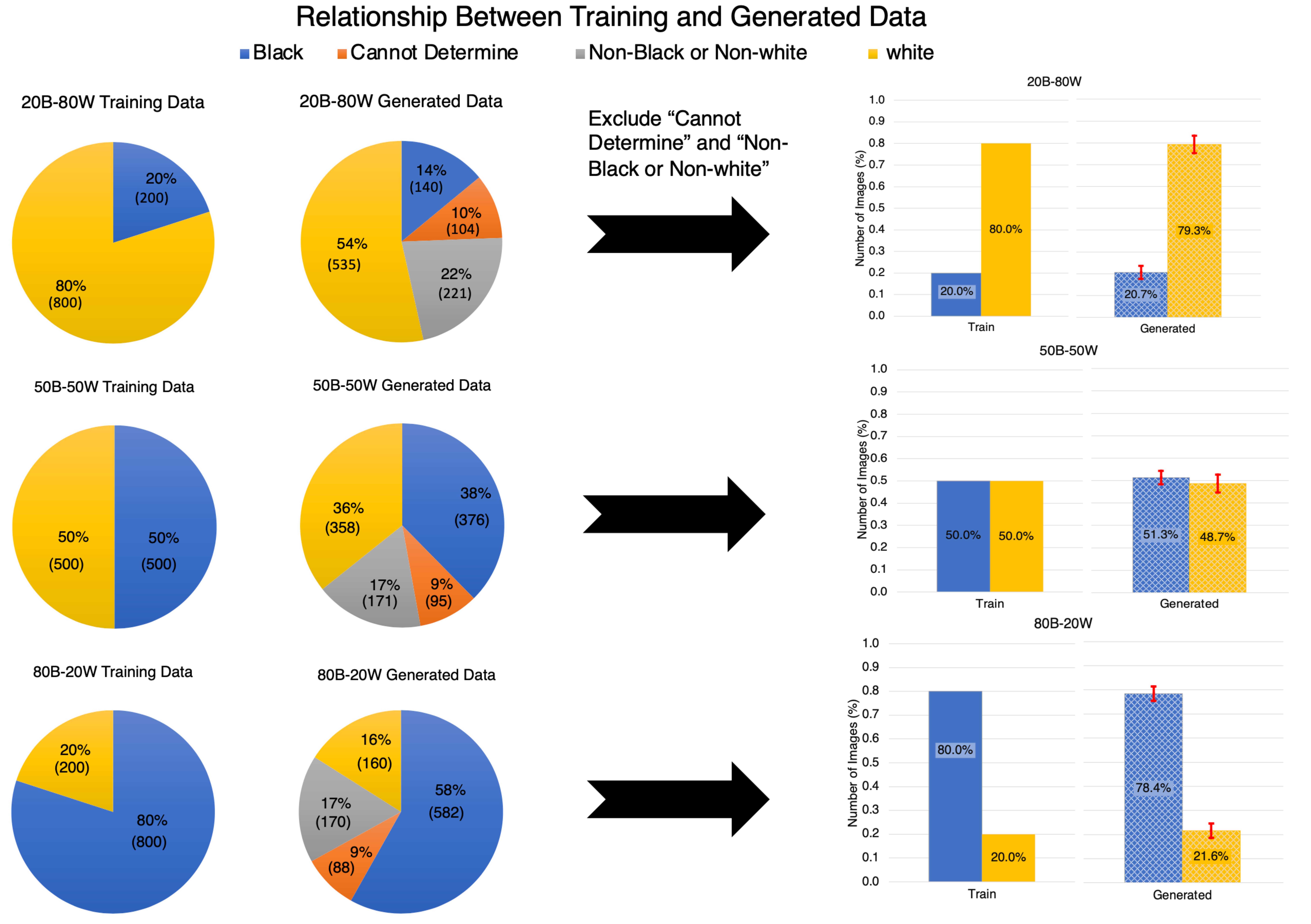} 
    \caption{Racial distribution of training and generated data. Distributions for  20B-80W, 50B-50W and 80B-20W data splits for (left) all the class labels and (right)  Black and white class labels only. This figure shows that all of the generative models preserve distribution of the training data.The red bars represents the 95\% Wald confidence interval (CI) of the generated data. See Table~\ref{tab:wald_CI_95} for the corresponding CI.}
    \label{fig:train_vs_gen}
\end{figure}

In this section, we expand on the results that demonstrate that StyleGAN2-ADA's generated data distribution preserves the training data distribution. In the paper, we excluded the ``Non-Black or Non-white” and ``Cannot Determine” class labels in the generated data to explicitly showcase the ratio of Black and white race class labels in the training and generated data. Fig.~\ref{fig:train_vs_gen} (left) shows the generated data distribution with all the class labels where the actual number of the classes are in bracket in the pie charts and Fig.~\ref{fig:train_vs_gen} (right) shows distribution for when the ``Non-Black or Non-white” and ``Cannot Determine” class labels were excluded. To get the generated data distribution with ``Non-Black or Non-white” and ``Cannot Determine” class labels, these two class labels were dropped and the distribution was recalculated with only white and Black class labels. 

\begin{table}[t]
    \centering
    \caption{Wald's 95\% confidence interval (CI) of generated data from generators trained on FFHQ and 80B-20W, 50B-50W and 80B-20W FairFace .} 

    \setlength{\tabcolsep}{12pt}
        \begin{tabular}{ccc}
            \toprule
             Generated Data & Black CI @95\% & white CI @95\% \\
            \midrule
FFHQ  & 5.73  $\pm$ 1.7   & 94.34 $\pm$ 1.7 \\
20B-80W & 20.70  $\pm$ 3.1   & 79.3 $\pm$ 3.1 \\
50B-50W  & 51.3  $\pm$ 3.7  & 48.7 $\pm$ 3.7 \\
80B-20W  & 78.4 $\pm$ 3.1  & 21.6 $\pm$ 3.1  \\

            \bottomrule
        \end{tabular}

    \label{tab:wald_CI_95}
\end{table}

\subsection{Truncation}

\begin{figure}[t]
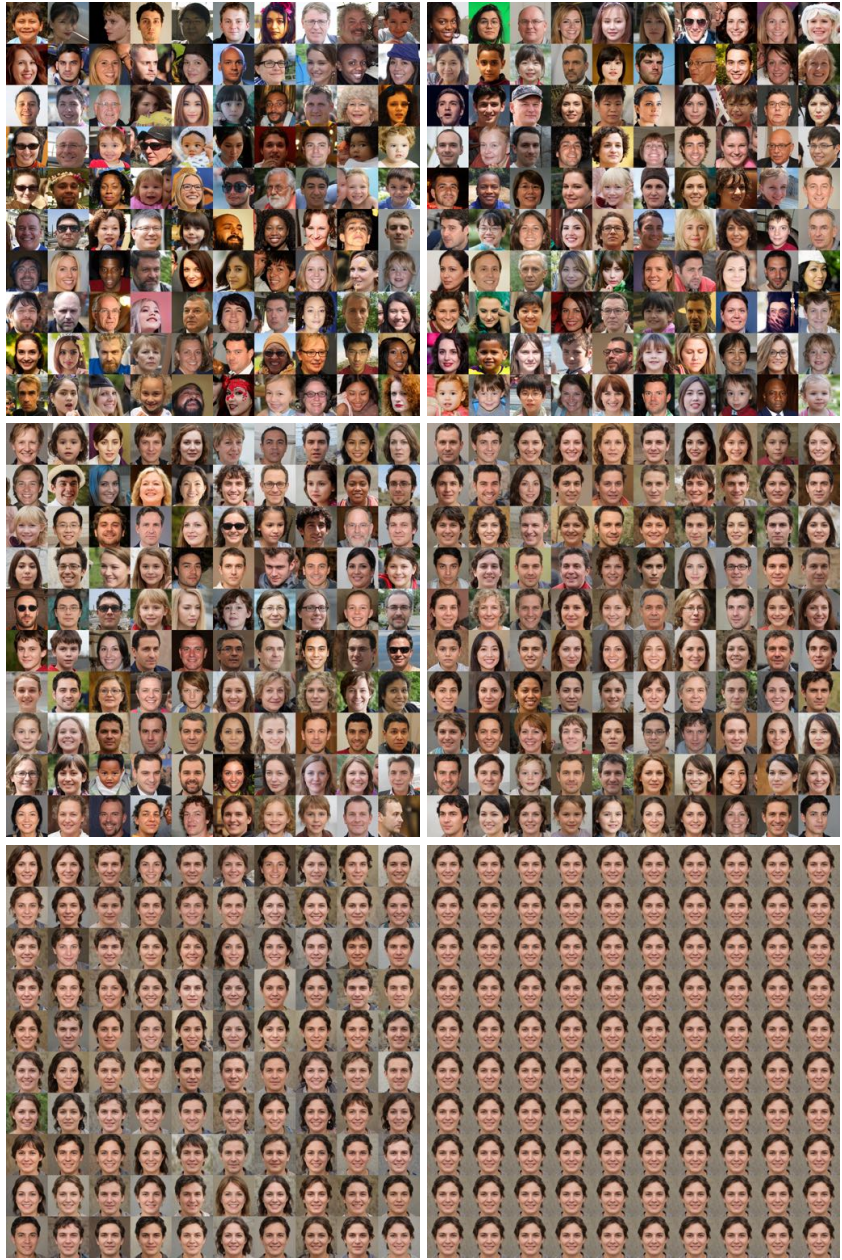


    \begin{center}
	\begin{tabular}{cc}
	 \includegraphics[width=0.45\textwidth]{figures/supp_truncation/ffhq_trunc_1.pdf} &  \includegraphics[width=0.45\textwidth]{figures/supp_truncation/ffhq_trunc_8.pdf}  \\
     \includegraphics[width=0.45\textwidth]{figures/supp_truncation/ffhq_trunc_6.pdf} &  \includegraphics[width=0.45\textwidth]{figures/supp_truncation/ffhq_trunc_4.pdf} \\
     \includegraphics[width=0.45\textwidth]{figures/supp_truncation/ffhq_trunc_2.pdf} &  \includegraphics[width=0.45\textwidth]{figures/supp_truncation/ffhq_trunc_0.pdf} 
    \end{tabular}
    \end{center}
    \caption{{\bf FFHQ samples with various levels of truncation.} (top left) truncation of $\gamma =1$ (no truncation), (top right) truncation of $\gamma = 0.8$, (middle left) truncation of $\gamma = 0.6$, (middle right) $\gamma =0.4$, (bottom left) truncation of $\gamma =0.2$, and (bottom right)  $\gamma =0$ (full truncation). As the amount of truncation increases, racial diversity decreases, resulting in an increasingly larger proportion of white faces.} 
    \label{fig:FFHQ-trunc-auto}
\end{figure}

In order to evaluate properties of truncation, images were generated from StyleGAN2-ADA trained on FFHQ and the 80B-20W, 50B-50W and 80B-20W FairFace-trained generators at truncation levels ranging from $\gamma=0$ to $1$ at intervals of $0.1$. Random samples from FFHQ with various levels of truncation can be seen in Fig.~\ref{fig:FFHQ-trunc-auto}. As the level of truncation increases, the ratio of faces of people of color to white faces decreases.

\subsection{Quality Ranking}

\begin{figure}[t]

    \begin{center}
	\begin{tabular}{cc}
	 \includegraphics[width=0.45\textwidth]{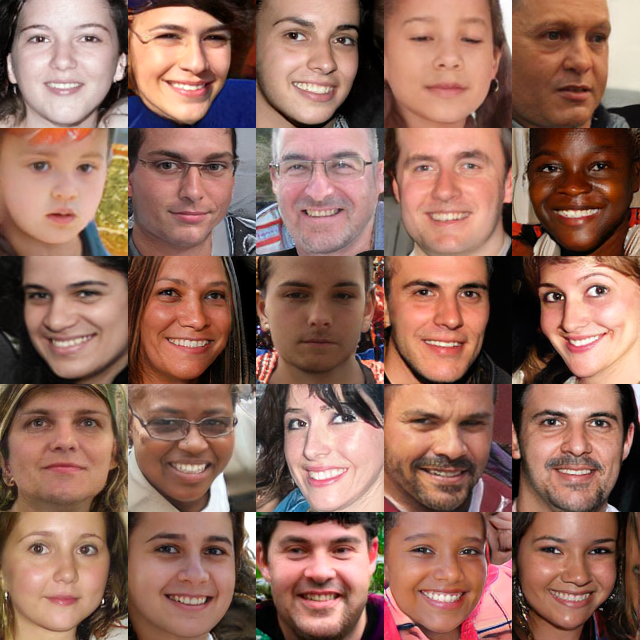} &  \includegraphics[width=0.45\textwidth]{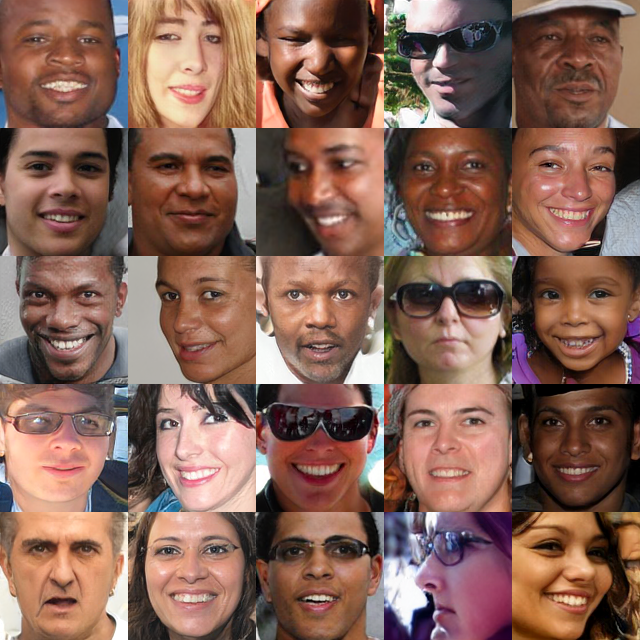}  \\
     \includegraphics[width=0.45\textwidth]{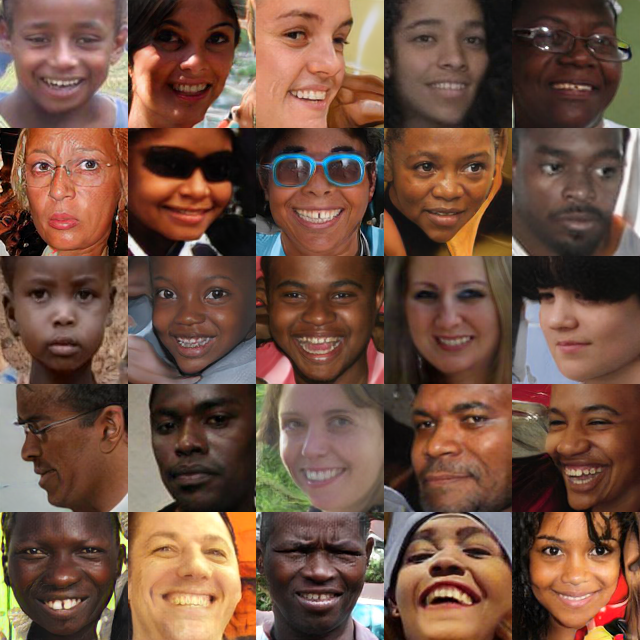} &  \includegraphics[width=0.45\textwidth]{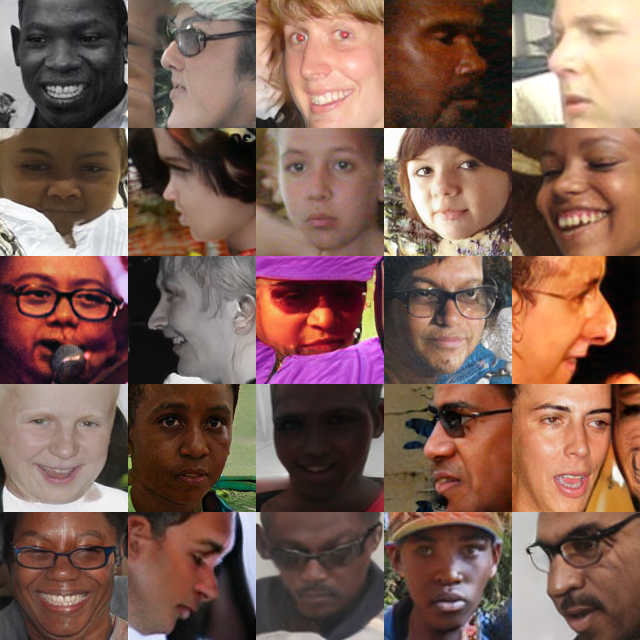} \\
     \includegraphics[width=0.45\textwidth]{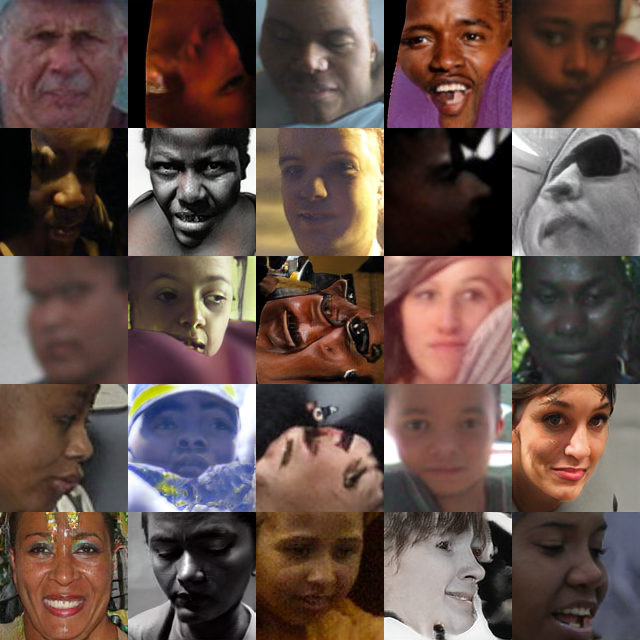} &  \includegraphics[width=0.45\textwidth]{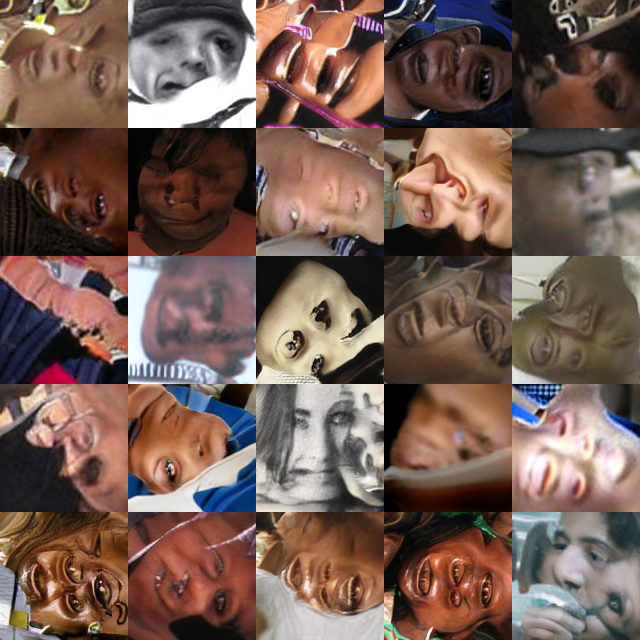} 
    \end{tabular}
    \end{center}
    \caption{{\bf Results of global quality ranking across the three FairFace data splits.} Each row represents a particular image quality ranking: (top left) top $25$ images in the quality ranking, (top right) samples from the top quartile of images, (middle left) samples from the second top quartile of images, (middle right) third quartile of ranked images, (bottom left) bottom quartile of ranked images, and (bottom right) the bottom $25$ images in the quality ranking. As shown in Table~\ref{tab:global-ranking-analysis}, white faces are over-represented in the top half of the quality ranking and under-represented in the bottom half.} 
    \label{fig:quality-quartiles}
\end{figure}

The raw intra-split pairwise comparison results from our quality ranking experiments can be seen in Tables~\ref{tab:generator-comparisons-ff-20-80}, \ref{tab:generator-comparisons-ff-50-50} and \ref{tab:generator-comparisons-ff-80-20}.
From these pairwise comparisons and the inter-split comparisons, we use the \texttt{choix} package~\cite{choix_git} to produce a Bradley-Terry model that ranks the $3000$ images in order of highest quality to lowest perceived image quality. From this global ranking, we visualize the top 25 and bottom 25 images, and also random images from each quartile, which can be seen in Fig.~\ref{fig:quality-quartiles}.

The raw counts for our top $K$ image compositions per race class label can be seen in Table~\ref{tab:raw-quality-race-breakdown}. In order to obtain weighted percentage scores seen in the main paper, the raw counts for the top $K$ images of a particular race class label were first weighted by the expected frequency of the images in each split. This was done by multiplying the raw count by $\frac{1}{2}$, $\frac{1}{5}$, and $\frac{1}{8}$ if the race class label comprised $20\%$, $50\%$ or $80\%$ of the dataset, respectively. Then, the weighted numbers were divided by the sum of all scores for that particular race class label and given value of $K$. 

A Bradley-Terry model~\cite{caron2012efficient} predicts the probability that a pairwise comparison $i > j$ is true. A ranking of all items can be derived by modeling the probability for pairs in a dataset. The \texttt{choix} package~\cite{choix_git} produces a Bradley-Terry model by using the Iterative Luce Spectral Ranking algorithm~\cite{maystre2015fast}. This algorithm performs maximum-likelihood inference to rank items from a dataset of pairwise comparisons.

\setlength{\tabcolsep}{4pt}
\begin{table}[t]
\begin{center}
\caption{{\bf Intra-split pairwise perceived image quality comparison 20B-80W Dataset.} }

    \resizebox{0.4\columnwidth}{!}{
        \begin{tabular}{lll}
            \hline
        \toprule
           More Preferred & Less Preferred & Count \\

            \hline
            \midrule

white        & white       & 2367  \\
white        & Black       & 708   \\
white        & Other       & 982   \\
white        & CD          & 1129  \\
Black        & Black       & 186   \\
Black        & white       & 465   \\
Black        & Other       & 205   \\
Black        & CD          & 270   \\
Other        & Other       & 411   \\
Other        & white       & 824   \\
Other        & Black       & 269   \\
Other        & CD          & 509   \\
CD           & CD          & 225   \\
CD           & Black       & 87    \\
CD           & white       & 230   \\
CD           & Other       & 133  \\
            \hline

            \bottomrule
        \end{tabular}
    }
    \label{tab:generator-comparisons-ff-20-80}
\end{center}
\end{table}

\setlength{\tabcolsep}{4pt}
\begin{table}[t]
\begin{center}
\caption{{\bf Intra-split pairwise perceived image quality comparison 50B-50W Dataset.} }

    \resizebox{0.4\columnwidth}{!}{
        \begin{tabular}{lll}
          \hline
        \toprule
            More Preferred & Less Preferred & Count \\

            \hline
            \midrule

white        & white       & 990   \\
white        & Black       & 1305  \\
white        & Other       & 497   \\
white        & CD          & 756   \\
Black        & Black       & 1167  \\
Black        & white       & 888   \\
Black        & Other       & 448   \\
Black        & CD          & 578   \\
Other        & Other       & 252   \\
Other        & white       & 475   \\
Other        & Black       & 599   \\
Other        & CD          & 323   \\
CD           & CD          & 159   \\
CD           & Black       & 274   \\
CD           & white       & 201   \\
CD           & Other       & 88   \\
            \hline

            \bottomrule
        \end{tabular}
    }
    \label{tab:generator-comparisons-ff-50-50}
\end{center}
\end{table}

\setlength{\tabcolsep}{4pt}
\begin{table}[t]
\begin{center}
\caption{{\bf Intra-split pairwise perceived image quality comparison of 80B-20W Dataset.} }

    \resizebox{0.4\columnwidth}{!}{
        \begin{tabular}{lll}
          \hline
        \toprule
            More Preferred & Less Preferred & Count \\

            \hline
            \midrule

white        & white       & 195   \\
white        & Black       & 798   \\
white        & Other       & 216   \\
white        & CD          & 319   \\
Black        & Black       & 2595  \\
Black        & white       & 660   \\
Black        & Other       & 705   \\
Black        & CD          & 1064  \\
Other        & Other       & 216   \\
Other        & white       & 216   \\
Other        & Black       & 918   \\
Other        & CD          & 295   \\
CD           & CD          & 219   \\
CD           & Black       & 421   \\
CD           & white       & 83    \\
CD           & Other       & 80  \\
\hline
            \bottomrule
        \end{tabular}
    }
    \label{tab:generator-comparisons-ff-80-20}
\end{center}
\end{table}

\setlength{\tabcolsep}{4pt}
\begin{table}[t]
\begin{center}
\caption{{\bf Race label breakdown of global ranking.} For each quartile of the global ranking of 3000 FairFace generated images compiled from different data splits, the percentage of faces annotated as white, Black, non Black/white, and Cannot Determine. white faces are over-represented in the top half of the quality ranking, and under-represented in the bottom half. }

    \resizebox{0.9\columnwidth}{!}{
        \begin{tabular}{ccccc}
        \hline
        \toprule
Quartile & white \% & Black \% & Non Black/white \% & Cannot Determine \% \\
            \hline
            \midrule

Top      & 48.4     & 26.3     & 22.5               & 2.8                 \\
Second   & 40.0     & 37.7     & 19.3               & 4.0                 \\
Third    & 29.7     & 41.3     & 17.2               & 11.7                \\
Bottom   & 15.2     & 32.1     & 11.6               & 41.1   \\
            \hline
        \bottomrule
        \end{tabular}
    }

    \label{tab:global-ranking-analysis}
    
\end{center}

\end{table}

\begin{table}[t]
    \centering

            \caption{{\bf Top $K$ image composition per-race, raw counts.} Given a ranking of Black and white images across all data splits, we break down the data split that each image came from. The highest quality images ($K=10, 25, 50$) are more likely to come from a data split where they are over-represented or represented in parity. These are the raw numbers, before normalization.}
        \begin{tabular}{cccc}
       \multicolumn{2}{c}{} & \multicolumn{1}{c}{\textbf{white}} & 
            \multicolumn{1}{c}{} \\ 
            \toprule
            $K$ & 80B-20W &  50B-50W & 20B-80W \\

            \midrule
            10  & 0 & 2 & 8 \\
            25  & 0& 7 & 18 \\
            50  & 4 & 16 & 30 \\
            100 & 10 & 39 & 51 \\
            500 & 68 & 181 & 251 \\
            \bottomrule
        \end{tabular}
        \qquad
        \begin{tabular}{cccc}
        \multicolumn{2}{c}{} & \multicolumn{1}{c}{\textbf{Black}} & \multicolumn{1}{c}{} \\
            \toprule
            $K$ & 80B-20W &  50B-50W & 20B-80W \\
       
            \midrule
            10  & 7 & 2& 1\\
            25  & 19 &4 &2\\
            50  & 31 & 16 &3\\
            100 & 57& 31 &12 \\
            500 & 275 & 161& 64 \\
            \bottomrule
        \end{tabular}
    
    \label{tab:raw-quality-race-breakdown}
\end{table}

\clearpage
%
%
\bibliographystyle{splncs04}
\bibliography{main}